\documentclass[10pt,twocolumn,letterpaper]{article}

\usepackage{cvpr}   
\usepackage{multirow} 
\usepackage{contour}
\usepackage{amsbsy}
\usepackage{subcaption}
\usepackage{xcolor} % 如需调整颜色可加，否则可省略
\newcommand{\thickhline}{\noalign{\hrule height 1pt}} 
\usepackage{amsmath, amssymb}
\usepackage{microtype}
\usepackage{colortbl}
\usepackage{array}
\usepackage{booktabs}
\usepackage{xcolor}
\usepackage{float}

\usepackage{arydshln}
\usepackage{makecell}
\definecolor{mygray}{gray}{0.9}
\usepackage[table]{xcolor}
\usepackage{float}
\definecolor{cvprblue}{rgb}{0.21,0.49,0.74}
\usepackage[pagebackref,breaklinks,colorlinks,allcolors=cvprblue]{hyperref}

\def\paperID{14738} % *** Enter the Paper ID here
\def\confName{CVPR}
\def\confYear{2026}

\title{Reframing Long-Tailed Learning via Loss Landscape Geometry}

\author{
    Shenghan Chen$^{1}$\thanks{Equal contribution. \\ \indent $^\dagger$Corresponding author: Xiankai Lu (luxiankai@sdu.edu.cn)} \quad 
    Yiming Liu$^{1}$\footnotemark[1] \quad 
    Yanzhen Wang$^{1}$ \quad 
    Yujia Wang$^{2}$ \quad 
    Xiankai Lu$^{1\dagger}$ \\[2mm]
    $^{1}$Shandong University \quad $^{2}$Zhejiang Sci-Tech University \\
}

\begin{document}
\maketitle

\begin{abstract}
Balancing performance trade-off on long-tail (LT) data distributions remains a long-standing challenge. In this paper, we posit that this dilemma stems from a phenomenon called ``tail performance degradation'' (the model tends to severely overfit on head classes while quickly forgetting tail classes) and pose a solution from a loss landscape perspective.  We observe that different classes possess divergent convergence points in the loss landscape. Besides, this divergence is aggravated when the model settles into sharp and non-robust minima, rather than a shared and flat solution that is beneficial for all classes. 
In light of this, we propose a continual learning inspired framework to prevent ``tail performance degradation''.  To avoid inefficient per-class parameter preservation, a Grouped Knowledge Preservation module is proposed to memorize group-specific convergence parameters, promoting convergence towards a shared solution. Concurrently, our framework integrates a Grouped Sharpness Aware module to seek flatter minima by explicitly addressing the geometry of the loss landscape.
Notably, our framework requires neither external training samples nor pre-trained models, facilitating the broad applicability. Extensive experiments on four benchmarks demonstrate significant performance gains over state-of-the-art methods. The code is available at:\url{https://gkp-gsa.github.io/}.
\end{abstract}

\section{Introduction}
\label{sec:intro}
Deep learning has achieved remarkable success across numerous computer vision tasks~\cite{bayoudh2024survey,kaushal2024computer,esteva2021deep}. This success is often predicated on the availability of large-scale, well-curated, and balanced datasets, such as MS-COCO~\cite{lin2014microsoft}. However, data encountered in real-world scenarios frequently exhibits a highly imbalanced or long-tailed distribution~\cite{yang2022survey,Jian_2025_ICCV}. Models trained on such datasets tend to develop a strong bias towards the data-abundant head classes, resulting in significantly degraded performance on the tail classes.

\begin{figure}[t]
  \centering
  \includegraphics[width=1\linewidth]{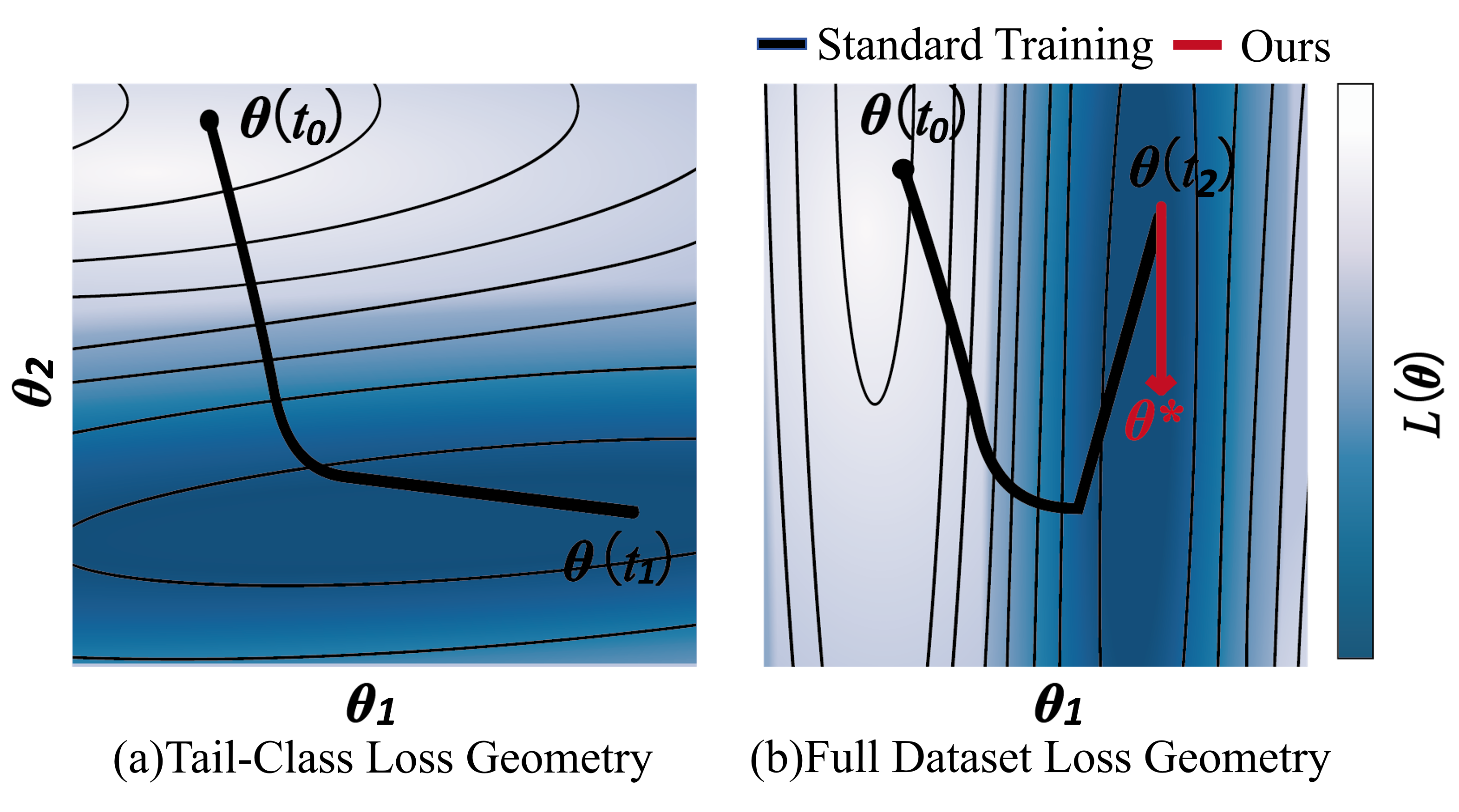} 
  \caption{``Tail performance degradation'' from the loss landscape view. Starting from a randomly initialized point $\theta(t_0)$ of LT model, (a) Training only on tail classes converges to $\theta(t_1)$ in a flat region, while (b) standard training on the long-tailed dataset converges to $\theta(t_2)$ in a sharp region. 
  The optimization trajectory settles in $\theta(t_2)$, which causes tail performance degradation by diverging from the tail convergence point $\theta(t_1)$. In contrast, our optimization (\textcolor{red}{red} line) steers the model towards a solution $\theta^*$ that remains closer to the tail-class minimum $\theta(t_1)$ and resides in a flatter region ($\theta_1$ and $\theta_2$ on the axes denote projection directions for 2D visualization~\cite{li2018visualizing}).}
  \label{fig:problem}
\end{figure}

A variety of approaches have been proposed to mitigate this challenge, broadly categorized into three groups~\cite{zhang2023deep}: class re-balancing~\cite{liu2008exploratory,cui2019class,hong2021disentangling}, information augmentation~\cite{wang2017learning,li2021metasaug} and module improvement~\cite{cui2021parametric,wu2020solving,zhou2020bbn} approaches. Despite their demonstrated effectiveness, these methods often encounter a fundamental trade-off dilemma~\cite{li2024long,Liu_2025_ICCV,Li_2025_ICCV}: enhancing the performance on tail classes leads to performance degradation on head classes, and vice versa (\ie, seesaw dilemma). While a recent trend involving more samples through external data~\cite{zhang2021mosaicos, zhang2023expanding} has proven to effectively address the dilemma, this approach is often infeasible in real-world scenarios (\eg, medical) where data is private. This limitation motivates a central research question: \textbf{What are the key factors in resolving the head-tail trade-off?} To answer this question, we delve  LT learning into the loss landscape view and analyze the optimization trajectories of model parameters during training~\cite{li2018visualizing,OptimizationTrajectories}.

In this work, we first visualize the optimization trajectories of a state-of-the-art long-tailed classifier, BCL~\cite{zhu2022balanced}, by projecting model parameters onto the loss landscape. Fig.~\ref{fig:problem} reveals two critical observations from the loss landscape view. First, the model suffers from ``tail performance degradation'': the standard optimization converges to $\theta(t_2)$ in Fig.~\ref{fig:problem} (b) that diverges significantly from the optimal convergence point for tail classes $\theta(t_1)$ in Fig.~\ref{fig:problem}(a)~\cite{foret2020sharpness}. This plot means the learned model focuses on head classes while rapidly forgetting the tail classes. Accordingly, the learned model with parameters $\theta(t_2)$ yields inferior performance for the tail classes than $\theta(t_1)$. 
Second, the model with standard training converges to a sharp minimum: compared to (a), the model settles at $\theta(t_2)$ (b) within a sharper region in the loss landscape. Accordingly, the model with $\theta(t_2)$ is inherently sensitive to the underlying perturbation and not robust for generalization across disparate classes. 

These two reasons make the current LT training paradigm fail to locate an optimal solution (\ie, $\theta^*$) yielding balanced recognition performance for both head and tail classes~\cite{li2024friendly}.

Although standard LT involves joint training without explicit task boundaries, dominant head-class gradients eventually pull the model from tail-friendly flat minima. Recognizing this implicit "forgetting", we formulate long-tailed learning as a continual learning (CL) task from the head classes to the tail classes. Thus, we transfer the head-tail balance issue into the \textit{knowledge preservation and acquisition} balance in continual learning. Building on the insights gained from our investigation, we propose a framework of preserving Knowledge and flattening landscapes that is composed of two branches: the \textit{Grouped Knowledge Preservation} (GKP) branch and the \textit{Grouped Sharpness Aware} (GSA) branch. The GKP branch mitigates tail performance degradation while GSA branch directs the optimization towards a flat convergence region, promoting convergence towards a unified solution beneficial for all classes. Subsequently, an adaptive parameter, scheduled according to the training epoch, is used to aggregate the losses from these branches.

% Therefore, to address the trade-off dilemma, we formulate long-tailed learning as a continual learning (CL) task from the head classes to the tail classes. Thus, we transfer the head-tail balance issue into the \textit{knowledge preservation and acquisition} balance in continual learning. Building on the insights gained from our investigation, we propose a framework of preserving Knowledge and flattening landscapes that is composed of two branches: the \textit{Grouped Knowledge Preservation} (GKP) branch and the \textit{Grouped Sharpness Aware} (GSA) branch. The GKP branch mitigates tail performance degradation while GSA branch directs the optimization towards a flat convergence region, promoting convergence towards a unified solution beneficial for all classes. Subsequently, an adaptive parameter, scheduled according to the training epoch, is used to aggregate the losses from these branches.

%This adaptive weighting mechanism guides the MKP model to acquire robust, generalizable features while preserving learned knowledge, ultimately steering the optimization towards a shared, flat solution.

The main contributions of this paper are as follows:
\begin{itemize}
\item We investigate the head-tail seesaw dilemma from the loss landscape view and posit the underlying factors, \ie, ``tail performance degradation'' and sharpness region.   

\item We transfer long-tailed recognition into a continual learning task to diagnose these issues. 

\item We propose a new long tail learning framework, including a Grouped Knowledge Preservation (GKP) branch to preserve existing knowledge and a Grouped Sharpness Aware (GSA) branch that flattens the loss landscape.
\item Extensive experiments show our method achieves state-of-the-art performance on four long-tailed visual recognition benchmarks.
\end{itemize}

\section{Related Work}
\label{sec:Related work}
\textbf{Long-tailed Learning: }Long-tailed recognition presents a significant challenge in computer vision, arising from the inherent imbalanced distribution of real-world data~\cite{cui2019class, zhu2024rectify, narayan2025segface, li2025conmix,hong2021disentangling}. Prevailing strategies to address this issue can be broadly categorized into three families: 1) Class Re-balancing~\cite{hong2021disentangling,zhang2021learning,liu2008exploratory,ren2020balanced}, which aims to counteract the optimization bias caused by class imbalance, typically through re-sampling or re-weighting; 2) Information Augmentation~\cite{shao2024diffult,li2021metasaug,liu2021gistnet,wei2021crest}, which enriches data-scarce classes by synthesizing or augmenting information; and 3) Module Improvement~\cite{huang2016learning,dong2017class,samuel2021distributional,wu2020solving}, which involves designing specialized network architectures or components to inherently better handle the class imbalance.

Recently, quite a few works have focused on leveraging external sources to incorporate additional training samples~\cite{zhang2021mosaicos,ramanathan2020dlwl,donglpt,shi2023parameter} or large pre-trained models~\cite {zhao2024ltgc,Song_2025_CVPR}. However, a significant limitation of this approach lies in the heavy reliance on external data or models. This requirement is often infeasible in practical scenarios where data privacy is paramount (\eg, medical applications). %Consequently, this constraint necessitates methods that operate without external data.  
Yet, these methods pay little attention to the underlying reason for the trade-off dilemma. This paper reframes  LT from the loss landscape view and proposes a new solution to alleviate the ``tail performance degradation''.   

%We posit that this stems from their failure to address the divergence between the optimal convergence points of different classes, and their corresponding lack of an effective mechanism to prevent tail performance degradation.

\smallskip
\noindent\textbf{Sharpness of Loss Landscape:} Research on model generalization increasingly focuses on loss landscape geometry. Extensive work has established that models converging to flatter minima exhibit superior generalization~\cite{keskar2017large,jiangfantastic,liu2022towards}. This principle underpins optimization methods like Sharpness-Aware Minimization (SAM)~\cite{foret2020sharpness} and Friendly Sharpness-Aware Minimization~\cite{li2024friendly}.

In long-tailed learning, SAM has been adapted to improve tail-class generalization. Early methods typically combined SAM with standard re-balancing techniques~\cite{rangwani2022escaping}. Recognizing different class requirements, subsequent works proposed more granular strategies, such as selectively applying SAM only to tail classes~\cite{zhou2023imbsam} or implementing fine-grained, per-class controls~\cite{zhou2023class,li2025focal}. These works have proven effective for imbalanced data. This work also extends the SAM framework for long-tailed classification by removing the head-dominated global perturbation direction to improve the performance of tail classes.

% Our Grouped Sharpness Aware (GSA) module operates at the group level, striking a deliberate balance between performance and computational cost. This design allows it to circumvent the performance limitations of a coarse head-tail split while avoiding the computational overhead associated with fine-grained per-class methods.

\begin{figure}
    \centering
    \includegraphics[width=1\linewidth]{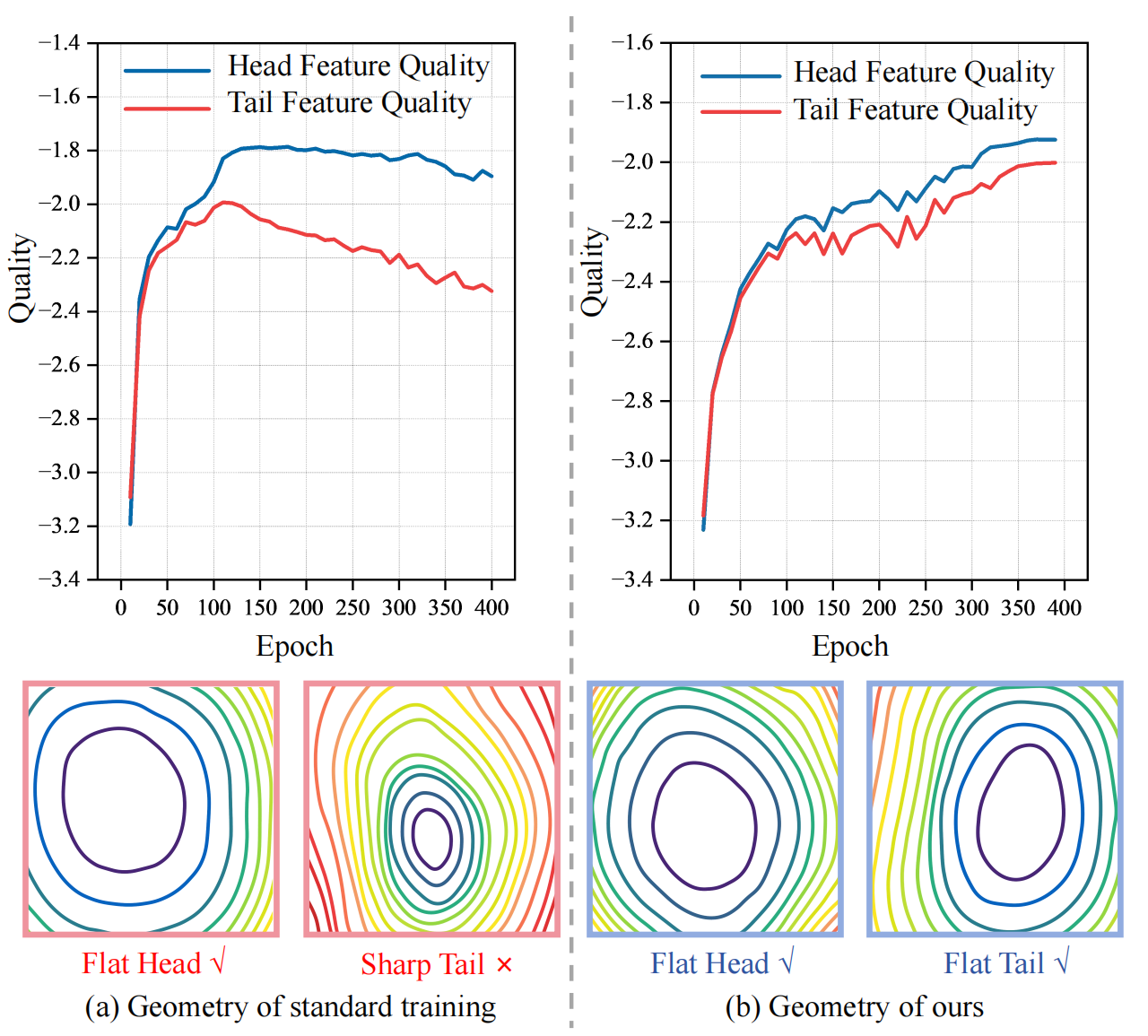}
    \caption{(a) Standard training results in a sharp loss landscape for tail classes, where the corresponding feature quality peaks and then declines. (b)  In contrast, our method flattens the landscape and preserves high feature quality for both head and tail classes.}
    \label{fig:exp}
\end{figure}

\section{The Feature Quality and Landscape: A Motivation Study}
\label{sec:Motivation}

\begin{figure*}[t] 
  \centering
    \includegraphics[width=0.85\textwidth]{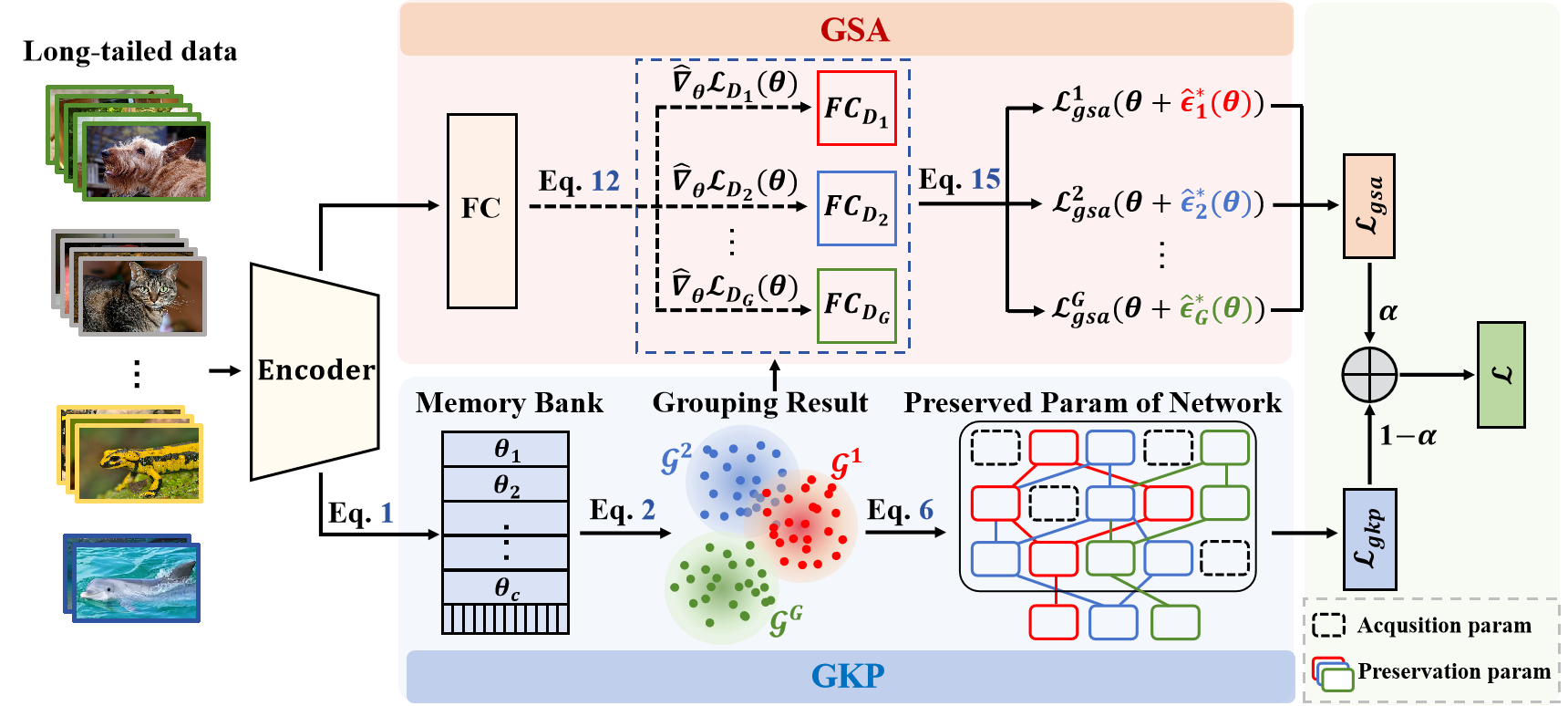} 
  \caption{Our framework consists of two key components: (a) The Grouped Sharpness Aware (GSA) module, which minimizes group-specific sharpness to find flat minima. (b) The Grouped Knowledge Preservation (GKP) module, which prevents tail performance degradation of other groups' optimal parameters.} %3) The Memory-based Grouping (MBG) strategy, which dynamically provides the shared class groups required by both GSA and GKP% 
  \label{fig:model}
\end{figure*}

%Our Introduction (Section \ref{sec:intro}) identifies tail performance degradation as parameter divergence from the tail-class convergence point. We posit that this parameter level phenomenon is a direct consequence of the optimization process being dominated by the feature representations of head classes. This section investigates this underlying feature level mechanism.

In this section, we conduct experiments from two dimensions: feature level and loss level to assess the influence of   ``tail performance degradation''.  All studies utilize BCL~\cite{zhu2022balanced} and our framework with CIFAR100-LT dataset~\cite{cao2019learning}. 

In the feature level, we define the \textit{Feature Quality} for head classes and tail classes \textbf{during} LT learning.   Feature Quality $Q$ is based on two components~\cite{majeescore}: inter-class separation, which measures the separation between different classes, and intra-class variance, which quantifies the dispersion of features within each class. In the loss level, we plot the loss landscape geometry for head classes and tail classes \textbf{after} model learning. Results are summarized in Fig.~\ref{fig:exp}.

\begin{itemize}
\item {\textbf{Observation 1}:
For BCL~\cite{zhu2022balanced}, the feature quality of the tail classes (red line) rises sharply in the initial phase, peaking at epoch 120, after which performance declines. While the feature quality of the head classes improves and maintains greater stability. For our framework, both head and tail classes exhibit a similar upward trend and stable enhancement in their feature quality.}

\textit{Analysis:} These results support our claim that ``the LT model tends to severely overfit on head classes while quickly forgetting tail classes with model training''. The superiority of our method confirms that knowledge preservation is a promising alternative.

\item{ \textbf{Observation} 2: For BCL, the tail classes have sharp loss landscape geometry, while our method has a flat region for both tail classes and head classes. } 

\textit{Analysis:} This provides direct evidence for our claim that current optimization ``is sensitive to the underlying perturbation and not robust for generalization across disparate classes''. The Grouped Sharpness Aware strategy, by design, controls the flatness of different classes to balance the performance between head and tail classes.
\end{itemize}

\section{Methodology}
\label{sec:Methodology}

\subsection{Overall Framework}
\label{sec:over}

This section presents our whole framework,  as illustrated in Fig.~\ref{fig:model}. Given a long-tailed training dataset $\mathcal{D}=\{ (x_i,y_i)^{N}_{i=1}\}$, where $x_i$ denotes a sample and $y_i\in \mathcal{C}$ 
represents its corresponding label with a total of $C$ classes. Our aim is to learn a function  $f_{\theta}$  with parameter $\theta$ mapping from an input space to the label space. The function $f_{\theta}$ is implemented as the composition of an encoder $a_i= f(\theta_{enc},x_i)$ with parameter $\theta_{enc}$ and a fully-connected
layer as classifier $W: x_i\rightarrow{\hat{y}_i}$. 

Different from most previous works~\cite{zhu2022balanced,li2021metasaug,Liu_2025_ICCV} that focus primarily on an effective encoder to enhance the long tail learning, this work rethinks the LT learning from a landscape view and devises a \textit{grouped knowledge preservation} module and a  \textit{grouped sharpness-aware} module to jointly ameliorate the encoder and the linear classifier.

 %$a_i=f(\theta_{enc}, x_i)$ denote the feature of $i$-th sample extracted by encoder with 

%During training, an input image with label $c$ and group index $g$ ($c \in \mathcal{V}_g$) is processed by a shared feature extractor, and the resulting features are fed into both the GSA and GKP branches simultaneously. 

\subsection{Grouped Knowledge Preservation Module} 
\noindent The design of the GKP module is inspired by the aim of continual learning (CL)~\cite{wang2024comprehensive,van2019three}, which addresses performance degradation: the tendency for knowledge of previously learned tasks (\eg, Task A) to be lost as information relevant to the current task (\eg, Task B) is incorporated. 

Unlike imbalanced Class-Incremental Learning (CIL), which tackles explicit sequential tasks with dual intra/inter-phase imbalances~\cite{he2024gradient}, long-tailed learning operates under a single joint objective. Thus, the key challenge in applying our CL-inspired scheme is how to define tasks. A naive per-class preservation strategy (treating each class as a task) is computationally prohibitive for datasets with many classes and severely hinders knowledge acquisition, as the optimization is constrained by the excessive number of preservation targets. Conversely, a simple head-tail task split is too coarse, ignoring the diverse convergence needs within the tail or head classes and thus failing to effectively preserve knowledge. Therefore, our GKP employs a memory-based grouping strategy to balance preservation and acquisition by clustering multiple classes based on their shared optimal parameters and subsequently treating each group as a task. 

% To prevent knowledge forgetting across different tasks (\eg, preserving tail classes knowledge) while maintain the knowledge acquisition for each specific task (\eg, effectively optimizing for the head classes), the GKP module first employs a memory-based grouping strategy to store the parameters for each class for the subsequent parameter preservation.   

% Considering preserving each class individually severely limits the model's plasticity for learning the head classes. To prevent knowledge forgetting across different tasks (\eg, preserving tail-class knowledge) and maintain the learning capability for each specific task (\eg, effectively optimizing for the head classes). Thus, we propose a memory based grouping strategy.

\subsubsection{Memory-based Grouping Strategy}
\label{sec:MBG}
\noindent\textbf{Memory Construction.} 
Memory-based Grouping Strategy first constructs a memory bank~\cite{Smith_2024_CVPR} $\mathcal{M}$ to dynamically store the encoder parameters $\theta_{enc}^c$ that achieved the highest feature quality for each class $c$ during training.

During model training, this memory bank is updated dynamically: at each epoch $t$, the current encoder parameters $\theta_{enc}^{(t)}$ replace the stored $\theta_{enc}^c$ if it yields a higher feature quality $Q$ (identified in Sec.~\ref{sec:Motivation}) for that class $c$:
\begin{equation}
\label{eq:updateBank}
\theta_{enc}^c \leftarrow \begin{cases}
    \theta_{enc}^{(t)} & \text{if } Q(\theta_{enc}^{(t)}, c) > Q(\theta_{enc}^c, c) \\
    \theta_{enc}^c & \text{otherwise}
\end{cases}
.
\end{equation}

Once the memory bank $\mathcal{M}$ is populated with the optimal encoder parameter sets $\mathcal{M}=\{{\theta_{enc}^1, \cdot \cdot \cdot , \theta_{enc}^C\}}$, we proceed to partition the class set $\mathcal{C} = \{1, \cdot \cdot \cdot , C\}$ into $G$ groups which would be detailed in the following section.

\noindent\textbf{Grouping Operation.}
Grouping Operation aims to group classes that exhibit similar encoder parameters. Based on $\mathcal{M}$, we leverage spectral clustering~\cite{ng2001spectral} method, \ie, Normalized Cuts (NCut) algorithm~\cite{shi2000normalized} to implement the operation:
\begin{equation}
\label{eq:ncut_solve}
\{\mathcal{G}^1,...\mathcal{G}^{g},...,\mathcal{G}^G\} = \text{NCut}(\mathcal{G}, G),
\end{equation}
where $G$ means the group number, $\mathcal{G}$ is a weighted, undirected graph $\mathcal{G} = (\mathcal{V}, \mathcal{E})$ which is built upon $\mathcal{M}$. $\mathcal{G}^{g}$ denotes the $g$-th sub-graph. During ablation studies (\S\ref{sec:ablation}), we perform experiments to assess the effect of group number. 

%first compute a $C \times C$ similarity matrix $\mathbf{A}$, where each element $\mathbf{A}_{i,j}$ quantifies the similarity between the optimal parameter vectors $\theta^i$ and $\theta^j$. Leveraging this matrix, we construct a weighted, undirected graph $\mathcal{G} = (\mathcal{V}, \mathcal{E})$, where the set of nodes $\mathcal{V}$ corresponds to the set of classes and the weights of the edges $\mathcal{E}$ are given by the similarity scores $\mathbf{A}$. The objective is to partition this graph into $G$ disjoint groups. Inspired by spectral clustering~\cite{ng2001spectral}, we employ the Normalized Cuts (NCut) algorithm~\cite{shi2000normalized} to find the partitioning $\mathcal{P} = \{\mathcal{V}_1, ..., \mathcal{V}_G\}$:

Once the group partitioning is obtained, we compute a shared encoder parameter $\theta_g^*$ for each group $\mathcal{G}^g$:  
\begin{equation}
\label{eq:group_optimum}
\theta_g^* = \frac{1}{|\mathcal{G}^g|} \sum_{c \in \mathcal{G}^g} \theta_{enc}^c.
\end{equation}

In this way, we can obtain the group-wise parameters, as illustrated in Fig.~\ref{fig:model}. 
%The resulting class partitioning $\mathcal{}$ and the set of shared optimal parameters $\Theta=\{\theta_1^*, ..., \theta_G^*\}$ are leveraged by GKP and GSA modules, enabling them to operate at the group level as illustrated in Fig.~\ref{fig:model}. 

\subsubsection{Parameter Preservation}
Regularization-based CL approaches focus primarily on mitigating tail performance degradation by penalizing changes to parameters vital for previously learned tasks. A typical solution is Elastic Weight Consolidation (EWC)~\cite{kirkpatrick2017overcoming}, which estimates the importance of each parameter using the Fisher Information Matrix as  corresponding quadratic penalty:
\begin{equation}
\label{eq:ewc_penalty}
\begin{aligned}
\mathcal{L}_{EWC} &= \mathcal{L}_{D}(\theta)+\mathcal{L}_{penalty}(\theta) \\
 &= \mathcal{L}_{D}(\theta) + \frac{\lambda}{2}\sum_i F_i (\theta_i - \theta_{t-1,i}^*)^2,
\end{aligned}
\end{equation}
where $\mathcal{L}_{D}$ works for knowledge acquisition while  $\mathcal{L}_{penalty}(\theta)$ works for knowledge preservation. 
$\lambda$ controls the strength of the regularization, $F_i$ is the diagonal of the Fisher Information Matrix, $\theta_{t-1}^*$ are the optimal parameters for previous tasks and $i$ denotes each parameter component.

%However, naively applying this strategy to long-tailed learning is suboptimal. We observe that different classes possess divergent convergence points. A standard EWC approach would either be computationally inefficient by protecting each class individually, or ineffective by using a coarse head-tail split that ignores the diverse needs of different class groups.

%Our Grouped Knowledge Preservation (GKP) module addresses the dilemma by operating in a more efficient group-wise manner, where we analogize ``tasks'' to the defined class groups (Eq.~\ref{eq:group_optimum}). 
Based on the group-wise parameters in Eq.~\ref{eq:group_optimum}, the penalty term $\mathcal{L}_{penalty}(\theta)$ in Eq.~\ref{eq:ewc_penalty} is reformulated as a dynamic, group-wise constraint. Specifically, when the model is training on the current group $g$ (\ie, knowledge acquisition), we revise $\mathcal{L}_{penalty}(\theta)$ to simultaneously preserve the knowledge of all other groups ($j \neq g$) (\ie, knowledge preservation):
\begin{equation}
\label{eq:regular}
\mathcal{L}_{penalty}(\theta) = \frac{\lambda}{2}\sum_i\sum_{j \neq g} F_{j,i} (\theta_i - \theta_{j,i}^*)^2,
\end{equation}
where $F_{j,i}$ denotes the $i$-th diagonal element of the (approximated) Fisher Information Matrix for group $j$~\cite{kirkpatrick2017overcoming}.
%Following standard CL methods, this penalty is weighted by the importance of each parameter $i$ to group $j$, . 

%Critically, $F_{j,i}$ values are naturally larger for groups with more samples. 
%these high-sample groups would dominate the total loss. To mitigate this problem and ensure all groups are also preserved equally, 
Considering the sample sizes for each group are various, we balance the importance of each group in Eq.~\ref{eq:regular} by normalizing group size $|\mathcal{G}^j|$ and obtain the final GKP loss:
\begin{equation}
\label{eq:gkp_penalty_term} 
\mathcal{L}_{gkp}^g = \frac{\lambda}{2} \sum_i \sum_{j \neq g} \left( \frac{1}{|\mathcal{G}^j|} F_{j,i} (\theta_i - \theta_{j, i}^*)^2 \right).
\end{equation}
 
We use the proposed GKP loss as the penalty term in the EWC loss function (Eq.~\ref{eq:ewc_penalty}) to implement knowledge preservation.  Meanwhile, for $\mathcal{L}_{D}$ in Eq.~\ref{eq:ewc_penalty}, we introduce a new optimization solution called grouped sharpness aware module to facilitate knowledge acquisition. 

\begin{figure}
    \centering
    \includegraphics[width=0.6\linewidth]{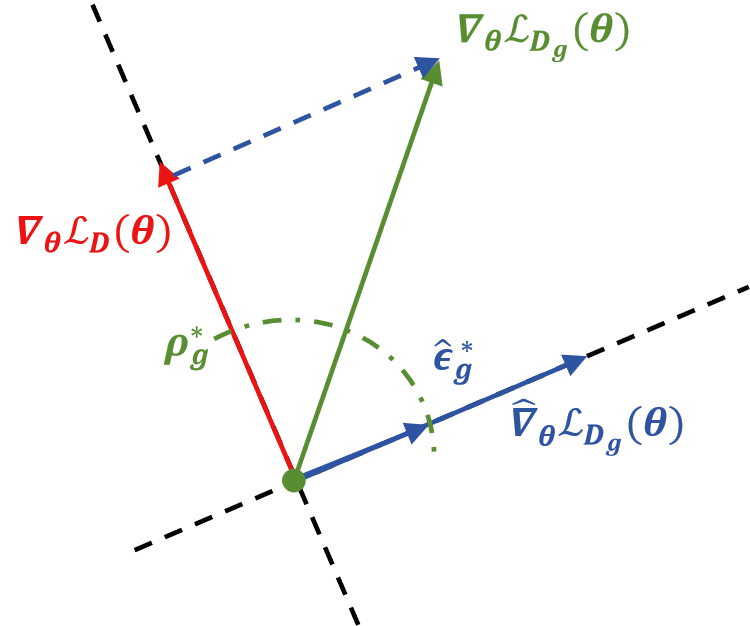}
    \caption{Investigation on the perturbation direction. We decompose the original gradient $\nabla_\theta \mathcal{L}_{\mathcal{D}_g}(\theta)$ (\textcolor[RGB]{88,142,49}{green}) into two components: the head-dominated global gradient $\nabla_\theta \mathcal{L}_D(\theta)$ (\textcolor[RGB]{236,19,19}{red}) and the beneficial, group-specific gradient $\hat{\nabla}_\theta \mathcal{L}_{\mathcal{D}_g}(\theta)$ (\textcolor[RGB]{46,84,161}{blue}).}
    \label{fig:perturbation}
\end{figure}

% Our full objective (Eq.~\ref{eq:gkp_full_split}) consists of two components. The first term $\log p(\mathcal{D}_g|\theta)$, corresponds to the standard loss function for the current group. The second term, which we define as our GKP loss. Taking the negative log-likelihood of the Gaussian approximation for each protected group (Eq.~\ref{eq:lap}) yields a tractable quadratic penalty. Meanwhile, to prevent groups with more samples from disproportionately influencing the parameter importance, we normalize this matrix by the group size  $n_j = |\mathcal{D}_j|$ for the group $j$. Thus the function that we minimize in GKP module is formulated as:
% \begin{equation}
% \label{eq:gkp_penalty_term} 
% \mathcal{L}_{gkp}^g = \frac{\lambda}{2} \sum_i \sum_{j \neq g} \left( \frac{1}{n_j} F_{j,i} (\theta_i - \theta_{j, i}^*)^2 \right),
% \end{equation}
% where $\lambda$ sets how important the protected groups is compared with the current group, and $i$ indicates each parameter in $\theta$.

\subsection{Grouped Sharpness Aware Module}
The Grouped Sharpness Aware (GSA) module is the second component of the framework, responsible for tailoring the knowledge acquisition objective $\mathcal{L}_{D}$ (Eq.~\ref{eq:ewc_penalty}) to obtain flat sharpness. Before elaborating our GSA, we first retrospect existing Sharpness-aware Minimization (SAM).

\noindent\textbf{Sharpness-aware Minimization.}
SAM is theoretically motivated by the PAC-Bayesian generalization bound~\cite{germain2016pac,reeb2018learning}, which establishes a link between a model's generalization error $\mathcal{L}_\mathcal{T}(\theta)$ and the sharpness of the loss landscape:
\begin{equation}
    \mathcal{L}_{\mathcal{T}}(\theta) \le \max_{||\epsilon||_2 \le \sqrt{d}\rho} \mathcal{L}_D(\theta+\epsilon) + \sqrt{\frac{\frac{||\theta||_2^2}{4\rho^2} + \log(\frac{N}{\delta}) + \mathcal{O}(1)}{N-1}},
\label{eq:pac_bound}
\end{equation}
where $\mathcal{L}_D(\cdot)$ comes from Eq.~\ref{eq:ewc_penalty}, $\mathcal{T}$ means the distribution of $D$. This bound holds for any perturbation radius $\rho > 0$ and $\delta \in (0, 1)$, with a probability of $1-\delta$. Moreover, $N$ represents the number of training samples, and $d := \text{dim}(\theta)$ denotes the dimensionality of the parameter space.

The objective of SAM is to find an optimal radius $\rho^*$ that minimizes this bound (Eq.~\ref{eq:pac_bound}): %Considering high noise ratios are known to degrade the classifier's decision boundary~\cite{zhou2023class,li2025focal}, we constrain our analysis to the pursuit of locally flat minima, which we define within the perturbative regime where $\rho \ll \|\theta\|_2$. 
%and optimal radius $\rho^*$:
\begin{equation}
\label{eq:max_rho}
    \rho^* \approx \left( \frac{\|\theta\|_2}{2\|\nabla_\theta \mathcal{L}_D(\theta)\|_2} \right)^{\frac{1}{2}} d^{-\frac{1}{4}} (N-1)^{-\frac{1}{4}},
\end{equation}
With the approximated optimal generalization bound, Eq.~\ref{eq:pac_bound} can be rewritten as:
\begin{equation}
\label{eq:final_bound}
    \hat{\mathcal{L}}_{\mathcal{T}}(\theta) \approx \max_{\|\epsilon\|_2 \le \sqrt{d}\rho^*} \mathcal{L}_{\mathcal{D}}(\theta + \epsilon) + \frac{1}{2\sqrt{N-1}}  \frac{\|\theta\|_2}{\rho^*}.
\end{equation}

The first term of this bound quantifies the sharpness of $\mathcal{L}_{\mathcal{D}}$, while the second term serves as a regularizer on the model parameters $\theta$. %Minimizing this approximated generalization bound $\hat{\mathcal{L}}_{\mathcal{T}}(\theta)$, is the objective for enhancing model generalization. % However, the presence of the maximization operator within the sharpness term renders the objective intractable for direct gradient-based optimization. This requires analytically solving the inner maximization problem. We 
The optimal perturbation vector $\hat{\epsilon}^*(\theta)$ can be computing as:
\begin{equation}
\label{eq:optimal_epsilon}
\begin{aligned}
    \hat{\epsilon}^*(\theta) &\approx \operatorname*{arg\,max}_{\|\epsilon\|_2 \le \sqrt{d}\rho^*} \left[ \mathcal{L}_{\mathcal{D}}(\theta) + \epsilon^T \nabla_\theta \mathcal{L}_{\mathcal{D}}(\theta) \right] \\
    &= \sqrt{d}\rho^* \frac{\nabla_\theta \mathcal{L}_{\mathcal{D}}(\theta)}{\|\nabla_\theta \mathcal{L}_{\mathcal{D}}(\theta)\|_2}.
\end{aligned}
\end{equation}

% However, the standard perturbation in Eq.~\ref{eq:optimal_epsilon} is dependent upon the global gradient, $\nabla_\theta \mathcal{L}_{\mathcal{D}}(\theta)$. In long-tailed datasets, this gradient is overwhelmingly dominated by head classes. Consequently, standard SAM becomes insensitive to the elevated sharpness associated with tail classes, resulting in a biased perturbation $\hat{\epsilon}^*$ that fails to adequately probe their sharp local landscapes. 
% To address this limitation, we propose the Group Sharpness-Awareness (GSA) module. Leveraging our MBG strategy (detailed in Sec 4.4), we partition classes into groups $\mathcal{V}_g$, where classes within each group exhibit similar convergence characteristics. This design ensures that while inter-group gradients may conflict, the intra-group gradient $\nabla \mathcal{L}_{\mathcal{D}_g}$ provides a stable and representative signal for its group's shared objective. 

\noindent\textbf{Grouped Sharpness-aware Minimization.}
However, the standard perturbation in Eq.~\ref{eq:optimal_epsilon} is dependent upon the global gradient $\nabla_\theta \mathcal{L}_{\mathcal{D}}(\theta)$, which is dominated by head classes. This renders standard SAM insensitive to the elevated sharpness of tail classes. To address this limitation, we propose the Grouped Sharpness-Awareness (GSA) module. 

Firstly, upon the grouping result in Eq.~\ref{eq:ncut_solve}, we shift from using the single global gradient (Eq.~\ref{eq:optimal_epsilon}) to calculating group-wise gradients $\nabla_\theta \mathcal{L}_{D_g}(\theta)$ for each group $ \mathcal{G}^g$:
\begin{equation}
\label{eq:gradient}
\nabla_\theta \mathcal{L}_{D_g}(\theta) \leftarrow [\nabla_\theta \mathcal{L}_{\mathcal{D}}(\theta), \mathcal{G}_g ].
\end{equation}

After that, considering our goal is to define a perturbation direction that is not biased by the head classes. Therefore, as shown in Fig.~\ref{fig:perturbation}, via the gradient decomposition analysis~\cite{li2024friendly}, 
we decompose $\nabla_\theta \mathcal{L}_{D_g}(\theta)$ (Eq.~\ref{eq:gradient}) into two components: 
\begin{equation}
\label{eq:gsa_direction}
\begin{split}
    \hat{\nabla}_\theta \mathcal{L}_{\mathcal{D}_g}(\theta) = & \nabla_\theta \mathcal{L}_{\mathcal{D}_g}{(\boldsymbol{\theta})}  - \operatorname{Proj}_{\nabla_\theta \mathcal{L}_{D}(\boldsymbol{\theta})}  \nabla_\theta \mathcal{L}_{\mathcal{D}_g}{(\boldsymbol{\theta})}, 
\end{split}
\end{equation}
where  $\nabla_\theta \mathcal{L}_{D}(\boldsymbol{\theta})$ is the global gradient over  $\mathcal{D}$ and $\operatorname{Proj}_{\nabla_\theta \mathcal{L}_{D}(\boldsymbol{\theta})}(\cdot)$ denotes the projection operator on $\nabla_\theta \mathcal{L}_{D_g}(\theta)$ along the direction of the global gradient.
In this way, we can obtain the group-specific gradient $\hat{\nabla}_\theta \mathcal{L}_{\mathcal{D}_g}(\theta)$.

Moreover,  to balance the perturbation radius across groups, we refine Eq.~\ref{eq:max_rho} by leveraging the group size $|\mathcal{G}^g|$:  % acknowledging these groups possess disparate sample sizes $n_g$, we introduce a size-dependent perturbation radius. This ensures that groups with fewer samples receive an appropriate degree of smoothing. We define $\rho_g^*$ to be inversely proportional to the group's sample size:
\begin{equation}
\label{eq:group_radius}
    \rho_g^* \approx \left( \frac{\|\theta\|_2}{2\|\hat{\nabla}_\theta \mathcal{L}_{\mathcal{D}_g}(\theta)\|_2} \right)^{\frac{1}{2}} d^{-\frac{1}{4}} (|\mathcal{G}^g|-1)^{-\frac{1}{4}}.
\end{equation}

By substituting this group-specific radius $\rho_g^*$ (Eq.~\ref{eq:group_radius}) and the corresponding group gradient  $\hat{\nabla}_\theta \mathcal{L}_{\mathcal{D}_g}(\theta)$ (Eq.~\ref{eq:gsa_direction}) into the optimal perturbation formula (Eq.~\ref{eq:optimal_epsilon}), we derive the final GSA perturbation vector $\hat{\epsilon}_g^*(\theta)$:
\begin{equation}
\label{eq:gsa_epsilon}
    \hat{\epsilon}_g^*(\theta) = \sqrt{d}\rho^*_g \frac{\hat{\nabla}_\theta \mathcal{L}_{\mathcal{D}_g}(\theta)}{\| \hat{\nabla}_\theta \mathcal{L}_{\mathcal{D}_g}(\theta)\|_2}.
\end{equation}

By using a group gradient (Eq.~\ref{eq:gsa_direction}) that removes the head-dominated global gradient, GSA improves tail classes flatness.
Consequently, the training objective of GSA module that works for knowledge acquisition is formulated as:
\begin{equation}
\label{eq:gsa_practical_loss}
\begin{aligned}
    \mathcal{L}_{gsa}^g(\theta) = \mathcal{L}_{\mathcal{D}_g}(\theta + \hat{\epsilon}_g^*(\theta)) + \frac{1}{2\sqrt{|\mathcal{G}^g|-1}} \frac{\|\theta\|_2}{\rho_g^*}.
\end{aligned}
\end{equation}

\subsection{Training Objects}
Our framework is jointly optimized by grouped sharpness-aware objective $\mathcal{L}_{gsa}^g$ (Eq.~\ref{eq:gsa_practical_loss})  and  grouped knowledge preservation objective $\mathcal{L}_{gkp}^g$ (Eq.~\ref{eq:gkp_penalty_term}):
\begin{equation}
    \label{eq:Loss}
         \mathcal{L} = \sum_{g=1}^G\left[\alpha \mathcal{L}_{gsa}^g + (1-\alpha) \mathcal{L}_{gkp}^g\right],
\end{equation}
where $\alpha$ is an adaptive parameter scheduled according to the training epoch~\cite{zhou2020bbn}.

\begin{table*}[ht]
    \caption{Results on CIFAR100-LT~\cite{cao2019learning} and CIFAR10-LT datasets~\cite{cao2019learning}. The imbalance ratio $r$ is set to 100, 50 and 10.  Additionally, we present the results for different groups (``Many'', ``Med.'' and ``Few'') in CIFAR100-LT with $r$ = 100. $^\dagger$ denotes methods that utilize Large Language Models.}
    \centering
    
    \resizebox{\textwidth}{!}{
    \begin{tabular}{|l|ccc|ccc|ccc|}
    % \toprule
    \hline\thickhline
    \rowcolor{mygray}
     & \multicolumn{3}{c|}{CIFAR100-LT} & \multicolumn{3}{c|}{CIFAR10-LT} & 
     \multicolumn{3}{c|}{Statistic($r$=100)} \\
    \rowcolor{mygray}
     \multirow{-2}{*} {Method} &  r=100$\uparrow$ & r=50$\uparrow$ & r=10$\uparrow$  & r=100$\uparrow$ & r=50$\uparrow$ & r=10$\uparrow$   & Many$\uparrow$ & Med.$\uparrow$ & Few$\uparrow$ \\ \hline\hline
    % \rowcolor{mygray}
    %\multicolumn{13}{|c|}{\textit{Close-World}} \\ 
    
         CE (\textcolor{gray}{Baseline}) & 38.3 & 43.9 & 55.7 & 70.4 & 74.8 & 86.4 & 65.2 & 37.1 & 9.1 \\
    Focal Loss~\cite{lin2017focal} (\textcolor{gray}{ICCV'17}) & 38.4 & 44.3 & 55.8 & 70.4 & 76.7 & 86.7 & 65.3 & 38.4 & 8.1 \\
    LDAM-DRW~\cite{cao2019learning} (\textcolor{gray}{NeurIPS'19})  & 42.0 & 46.6 & 58.7 & 77.0 & 81.0 & 88.2 & 61.5 & 41.7 & 20.2 \\
    cRT~\cite{kang2019decoupling} (\textcolor{gray}{ICLR'20})  & 42.3 & 46.8 & 58.1 & 75.7 & 80.4 & 88.3 & 64.0 & 44.8 & 18.1 \\   
    BBN~\cite{zhou2020bbn} (\textcolor{gray}{CVPR'20})  & 42.6 & 47.0 & 59.1 & 79.8 & 82.2 & 88.3 & - & - & - \\
    RIDE (3 experts)~\cite{wang2020long} (\textcolor{gray}{ICLR'21})  & 48.0 & - & - & - & - & - & 68.1 & 49.2 & 23.9 \\
    CAM-BS~\cite{zhang2021bag} (\textcolor{gray}{AAAI'21})  & 41.7 & 46.0 & - & 75.4 & 81.4 & - & - & - & - \\
    DiVE~\cite{he2021distilling} (\textcolor{gray}{ICCV'21})  & 45.4 & 51.1 & 62.0 & - & - & - & - & - & - \\
    SAM~\cite{rangwani2022escaping} (\textcolor{gray}{NeurIPS'22})  & 45.4 & - & - & 81.9 & - & - & 64.4 & 46.2 & 20.8 \\
     BCL~\cite{zhu2022balanced} (\textcolor{gray}{CVPR'22}) & 51.9  & 56.6 & 64.9 & 84.3& 87.2 & 91.1 & 67.2 & 53.1 & 32.9 \\
    CUDA~\cite{ahn2023cuda} (\textcolor{gray}{ICLR'23})  & 47.6 & 51.1 & 58.4 & - & - & - & 67.3 & 50.4 & 21.4 \\
    ADRW~\cite{wang2023unified} (\textcolor{gray}{NeurIPS'23})  & 46.4 & - & 61.9 & 83.6 & - & 90.3 & - & - & - \\
    H2T~\cite{li2024feature} (\textcolor{gray}{AAAI'24}) & 48.9 & 53.8 & - & - & - & - & - & - & - \\
    GBG~\cite{li2024long} (\textcolor{gray}{AAAI'24}) & 52.3 & 57.2 & - & 85.1 & 87.7 & - & - & - & - \\
    DiffuLT~\cite{shao2024diffult} (\textcolor{gray}{NeurIPS'24})  & 51.5 & 56.3 & 63.8 & 84.7 & 86.9 & 90.7 & 69.0 & 51.6 & 29.7 \\
    DiffuLT + BBN~\cite{shao2024diffult} (\textcolor{gray}{NeurIPS'24})  & 51.9 & 56.7 & 64.0 & 85.0 & 87.2 & 90.9 & \textbf{69.5} & 51.9 & 30.2 \\
  
     SEL~\cite{Jian_2025_ICCV} (\textcolor{gray}{ICCV'25})  & 52.3 & 57.3 &  68.4 & 84.4 & 86.3 & 90.2 & - & - & - \\
     Heuristic-CALA\cite{zhou2025class} (\textcolor{gray}{AAAI'25})  & 50.5 & - &  64.3 & 83.9 & - & 91.7 & - & - & -\\
     Meta-CALA\cite{zhou2025class} (\textcolor{gray}{AAAI'25})  & 52.3 & - &  65.5 & 84.7 & - & 92.4 & - & - & -
     \\
       FeatRecon  \cite{yi2025geometry} (\textcolor{gray}{ICLR'25})  & 52.5 & 57.0 &  65.3 & 85.2 & 87.8 & 91.6 & - & - & -
     \\
     LLM-AutoDA$^{\dagger}$~\cite{DBLP:conf/nips/Wang0WWW0024}(\textcolor{gray}{NeurIPS'24})    & 51.0 & 54.8 & - & - & - & - & 66.6 & 50.6 & 33.1 \\
         \hdashline
       \textbf{Ours} & \textbf{53.2} & \textbf{57.6} & \textbf{68.7} & \textbf{86.3} & \textbf{88.2} & \textbf{92.5} & 67.3 & \textbf{54.9} & \textbf{34.9} \\

    \hline\thickhline
        \end{tabular}}
    
    \label{tab:cifar}
\end{table*}

\section{Experiment}

\subsection{Experiment Setup}

\textbf{Datasets.} Our proposed framework is evaluated on four long-tailed benchmarks:      CIFAR10-LT~\cite{cao2019learning}, CIFAR100-LT~\cite{cao2019learning}, ImageNet-LT~\cite{liu2019large} and iNaturalist 2018~\cite{van2018inaturalist}.

\noindent\textbf{Metrics.} Following standard evaluation protocols, we report Top-1 accuracy for comparison with state-of-the-art methods. To evaluate our method's effectiveness across varying imbalance levels, we report Top-1 accuracy under three imbalance ratios $r \in \{100,50, 10\}$ for CIFAR10-LT and CIFAR100-LT. Additionally, we report results for ``Many'' (classes with over 100 samples), ``Med.'' (classes with 20 to 
100 samples), and ``Few'' (classes with fewer than 20 samples) categories separately to enable in-depth analysis, following the methodology
described in~\cite{cao2019learning}. For ImageNet-LT, we present results with different feature backbones for a comprehensive evaluation of our method.

\noindent\textbf{Implementation Details.} For both CIFAR10-LT and CIFAR100-LT, we use the ResNet-32 as the backbone following~\cite{cao2019learning}.  For ImageNet-LT datasets, we
use ResNet-50~\cite{he2016deep} and ResNeXt50~\cite{xie2017aggregated} as backbone. For iNaturalist 2018, we use ResNet-50~\cite{he2016deep} as backbone. For all datasets, we set the batch size to 256 and train all models on NVIDIA GeForce RTX 3090 GPU. For further
implementation details, please refer to the Appendix. 
\subsection{Main Results}
\noindent\textbf{ CIFAR10-LT and CIFAR100-LT}~\cite{cao2019learning}. The comparison results between the proposed method and other existing methods on long-tailed
CIFAR are shown in Table ~\ref{tab:cifar}. Our proposed method surpasses recent state-of-the-art approaches across the datasets under various imbalance ratios.  On CIFAR100-LT, our method surpasses competing models, achieving accuracy improvements of 14.9\%, 13.7\%, and 13.0\% compared with the
 baseline for $r$ = 100, 50, and 10, respectively.  On CIFAR10-LT, our model also demonstrates strong competitiveness, enhancing accuracy by 15.9\%, 13.4\%, and 6.1\% for $r$ = 100, 50, and 10, respectively, further validating the effectiveness of our method.  
    
As the main counterpart, our method yields better performance than BCL~\cite{zhu2022balanced} (CIFAR100-LT: 53.2 vs. 51.9, 57.6 vs. 56.6, 68.7 vs. 64.9; CIFAR10-LT: 86.3 vs. 84.3, 88.2 vs. 87.2, 92.5 vs. 91.1). Considering both methods use the same backbone and loss functions, we attribute the performance improvement solely to the proposed grouped knowledge preservation module and the grouped sharpness aware module. 
 
For CIFAR100-LT with an imbalanced ratio of 100, we also assess performance across three categories: (``Many'': 67.3\%, ``Med.'': 54.9\%, ``Few'': 34.9\%), effectively addressing the performance trade-off between head and tail classes in long-tailed learning. These results collectively demonstrate our method's effectiveness in handling the fundamental challenges of long-tailed distributions, especially the problem of extreme class imbalance.

Notably,  our proposed method also surpasses the LLM-based LLM-AutoDA~\cite{DBLP:conf/nips/Wang0WWW0024} by 2.2\% in performance, despite using only the standard training set without any external data or pre-trained models. %This data-efficient advantage verifies the powerful role of our proposed modules.

\begin{table}[t] 
  \centering
%  \footnotesize % 使用小字体
\small
  \setlength{\tabcolsep}{3pt} % 减小列间距
    \caption{ Top-1 accuracy of ResNet-50  on ImageNet-LT~\cite{liu2019large} and iNaturalist 2018~\cite{van2018inaturalist}. $^\dagger$ denotes methods that utilize Large Language Models.  }

  \begin{tabular}{|l|c|c|}

    \hline\thickhline
    
    % --- 修改: 添加 \rowcolor 并使用 \hline ---
    \rowcolor{mygray} 
    Method & ImageNet-LT & iNature2018 \\
    \hline\hline
    
    % --- 修改: 将 \midrule 替换为 \hline ---
    CE(\textcolor{gray}{Baseline})& 41.6 & 61.0 \\
    Focal Loss~\cite{lin2017focal}(\textcolor{gray}{ICCV'17})  & - & 61.1 \\
    cRT~\cite{kang2019decoupling}(\textcolor{gray}{ICLR'20})  & 47.3 & 65.2 \\

    RIDE (3 experts)~\cite{wang2020long}(\textcolor{gray}{ICLR'21})  & 54.9 & 72.7 \\
    BCL~\cite{zhu2022balanced} (\textcolor{gray}{CVPR'22})& 56.0 & 71.8 \\
    SAM~\cite{rangwani2022escaping}(\textcolor{gray}{NeurIPS'22})  & 53.1 & 70.1 \\
    CUDA~\cite{ahn2023cuda}(\textcolor{gray}{ICLR'23}) & 51.4 & 72.2 \\
    ADRW~\cite{wang2023unified}(\textcolor{gray}{NeurIPS'23})  & 54.1 & 70.7 \\
    
     GBG~\cite{li2024long} (\textcolor{gray}{AAAI'24})& 57.6 & 71.9 \\
    DiffuLT~\cite{shao2024diffult}(\textcolor{gray}{NeurIPS'24})  & 56.4 & - \\
    DiffuLT + RIDE (3 experts)~\cite{shao2024diffult}& 56.9 & - \\
    (\textcolor{gray}{NeurIPS'24})&& \\ 
    SEL~\cite{Jian_2025_ICCV} (\textcolor{gray}{ICCV'25})  &56.3 &71.3 \\
    Heuristic-CALA~\cite{zhou2025class}  (\textcolor{gray}{AAAI'25}) &54.1 &73.2 \\  
    Meta-CALA~\cite{zhou2025class}  (\textcolor{gray}{AAAI'25}) &55.1 &74.0 \\  
    FeatRecon~\cite{yi2025geometry} (\textcolor{gray}{ICLR'25})  &56.8 &72.9 \\
    LLM-AutoDA$^{\dagger}$~\cite{DBLP:conf/nips/Wang0WWW0024}   (\textcolor{gray}{NeurIPS'24})& 57.5 & 74.2 \\
     \hdashline
    \textbf{Ours} & \textbf{57.9} & \textbf{74.4} \\
    \hline\thickhline
  \end{tabular}
  \label{tab:imagenet_inature} 
\end{table}
\noindent\textbf{ImageNet-LT}~\cite{liu2019large}. Table ~\ref{tab:imagenet_inature} and Table ~\ref{tab:resnext50}   report the results on
ImageNet-LT with different backbones for comprehensive results comparison.  We report the overall Top-1 accuracy as well  as the Top-1 accuracy on ``Many'', ``Medium'', and
 ``Few'' groups. Utilizing ResNet-50 backbone (Table ~\ref{tab:imagenet_inature}), our method registers 57.9\% accuracy, outperforming the state-of-the-art LLM-AutoDA~\cite{DBLP:conf/nips/Wang0WWW0024} by 0.4\%. 
 
With the stronger ResNeXt-50 backbone (Table ~\ref{tab:resnext50}), the accuracy of our method further escalates to 58.9\%. Specifically, our method achieves 68.7\%, 56.8\%, and 38.6\% on the ``Many'', ``Medium'', and ``Few'' splits,  outperforms the top-leading FeatRecon~\cite{yi2025geometry} by 1.8\%, 1.3\%, and 0.8\%, respectively. These significant gains, particularly on tail classes, highlight the effectiveness of our proposed grouped knowledge preservation module and the grouped sharpness aware module in addressing extreme data distribution imbalances. Again, these gains come from using only standard training set without any external data.
 \begin{table}[t]

    \caption{Top-1 accuracy of ResNeXt-50 on ImageNet-LT~\cite{liu2019large}.}
    \centering
    
    \resizebox{0.47\textwidth}{!}{
    \begin{tabular}{|l|cccc|}
    \hline\thickhline

    \rowcolor{mygray}
    Method & Many$\uparrow$ & Med.$\uparrow$ & Few$\uparrow$ & All$\uparrow$ \\ 
    
    \hline \hline
     CE (\textcolor{gray}{Baseline}) & - & - & - & 44.4 \\
    FocalLoss~\cite{lin2017focal} (\textcolor{gray}{ICCV'17}) & 64.3 & 37.1 & 8.2 & 43.7 \\
    $\tau$-norm~\cite{kang2019decoupling} (\textcolor{gray}{ICLR'20}) & 59.1 & 46.9 & 30.7 & 49.4 \\
    BalancedSoftmax~\cite{ren2020balanced} & 62.2 & 48.8 & 29.8 & 51.4 \\
    (\textcolor{gray}{NeurIPS'20}) &&&&\\
     Casualmodel~\cite{tang2020long}  & 62.7 & 48.8 & 31.6 & 51.8 \\(\textcolor{gray}{NeurIPS'20})&&&&\\
    
    LWS~\cite{kang2019decoupling} (\textcolor{gray}{ICLR'20}) & 60.2 & 47.2 & 30.3 & 49.9 \\
    LADE~\cite{hong2021disentangling} (\textcolor{gray}{CVPR'21}) & 62.3 & 49.3 & 31.2 & 51.9 \\
   DisAlign~\cite{zhang2021distribution} (\textcolor{gray}{CVPR'21}) & 62.7 & 52.1 & 31.4 & 53.4 \\
    RIDE (2 experts)~\cite{wang2020long}  & - & - & - & 55.9 \\(\textcolor{gray}{ICLR'21})&&&&\\
    BCL~\cite{zhu2022balanced} (\textcolor{gray}{CVPR'22}) & - & - & - & 56.7 \\
    % FDC~\cite{} (\textcolor{gray}{CVPR'22}) & 65.5 & 51.9 & 37.8 & 55.3 \\
    % IWB~\cite{dang2024inverse} (\textcolor{gray}{AAAI'24}) & 64.2 & 52.2 & \textbf{40.2} & 55.2\\
    GBG~\cite{li2024long} (\textcolor{gray}{AAAI'24}) & 69.6 & 55.8 & 38.1 & 58.7\\
   FeatRecon~\cite{yi2025geometry} (\textcolor{gray}{ICLR'25}) & 67.9 & 54.7 & 37.8 & 57.5 \\
    \hdashline 
    \textbf{Ours} & \textbf{69.7} & \textbf{56.0} & \textbf{38.6} & \textbf{58.9} \\
    
    % --- 样式清理 ---
    \hline \thickhline
    \end{tabular}}
    \label{tab:resnext50}       
\end{table}

\noindent\textbf{iNaturalist 2018} ~\cite{van2018inaturalist}.
Table~\ref{tab:imagenet_inature} (right column)  shows the experimental results on the real-world large-scale imbalanced iNaturalist 2018. Employing a ResNet-50 backbone, our method achieves an accuracy of 74.4\%, demonstrating a significant performance promotion of 13.4\% over the baseline.  %and surpassing the current state-of-the-art method by 0.2\%.

All of the above improvement on imbalanced recognition confirms the effectiveness of our proposed grouped knowledge preservation and grouped sharpness aware module which learns in a compositional manner, and is informed by parameter preservation and flatten landscape to effectively address ``tail performance degradation''. %Notably, all methods without using external data, the performance promotion is our method  %yielded only minor performance improvements. Our performance gain is relatively high by comparison.

\subsection{Ablation Study}
\label{sec:ablation}
In this section, we perform several ablation studies to characterize the
proposed the method. %All experiments are performed on
%CIFAR-100 with an imbalance factor of $r$=100, employing ResNet-32 as the classifier backbone.
    
\noindent\textbf{Component Analysis.} 
We first verify the effectiveness of
main components of our proposed framework. We specifically choose CIFAR100-LT ($r$=100) for experiment evaluation and use ResNet-32 as the backbone. 

As shown in Table ~\ref{tab:Ablation_components}, using  GKP modules brings +3.7\%  accuracy on \textit{CE} baseline. Furthermore, combining GKP and GSA modules then achieves a total 4.4\% improvement. On the stronger BCL~\cite{zhu2022balanced} baseline, our GSA and GKP modules yield  0.8\% and 0.5\% gain, respectively, while the full model achieves a 1.3\% total improvement, finally reaching 53.2\%. These results confirm the crucial and highly additive contribution of each component. 
    
%Furthermore, these gains are consistently observed across all class groups; compared to the BCL baseline, our full model improves accuracy on head classes by 1.0\%, medium classes by 1.3\%, and tail classes by 1.4\%, respectively.

\begin{figure}
    \centering
    \includegraphics[width=0.9\linewidth]{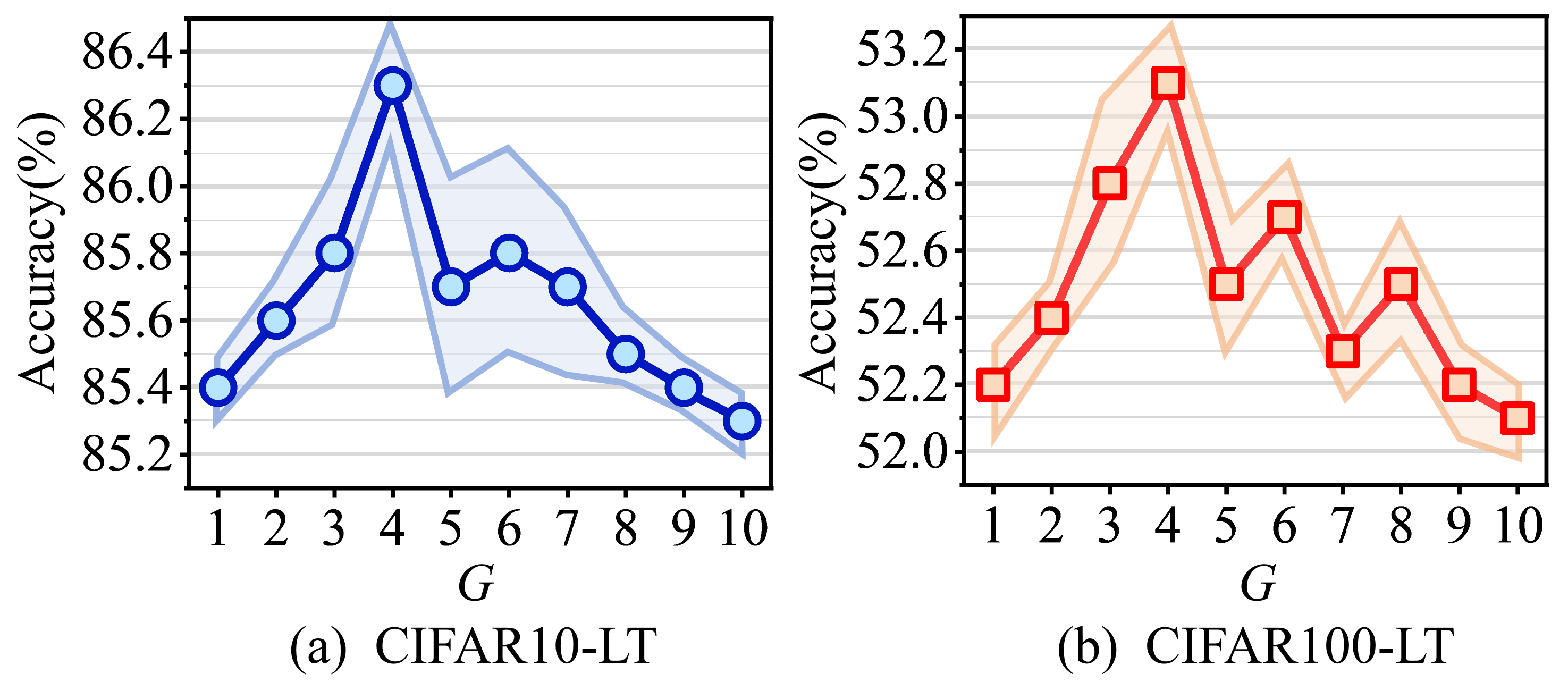}
    \caption{Ablation Study on group numbers of our method. The shaded area indicates the fluctuation of accuracy.}
    \label{fig:ablation}
\end{figure}

\noindent\textbf{Group Numbers.} 
In Fig.~\ref{fig:ablation}, we investigate the impact of different group numbers within our grouping strategy (\ie, $G$ in Eq.~\ref{eq:ncut_solve}). This study was conducted on CIFAR10-LT (\textcolor[RGB]{0,24,192}{blue}, left) and CIFAR100-LT (\textcolor[RGB]{255,0,0}{red}, right), both with an imbalance factor $r=100$. The results reveal that $G=4$ groups yield optimal performance as employing an excess number of groups does not necessarily enhance performance in the GKP modules. We report the detailed ablation studies for $G$ on other datasets  in the Appendix.
\begin{table}[ht]
    \caption{Ablation study on main components of our method.}
    \centering
    \small % <-- 使用 \small 字体
    % \resizebox{0.48\textwidth}{!}{ % <-- 已移除
    \begin{tabular}{|l|cccc|}
    \hline\thickhline
    \rowcolor{mygray}
    Method & Many$\uparrow$ & Med.$\uparrow$ & Few$\uparrow$ & All$\uparrow$ \\ 
    \hline\hline % <-- 修正了 \hline \hline
     CE & 65.2 & 37.1 & 9.1 & 38.4\\
    +GKP & 71.1 & 41.2 & 9.3 & 42.1 \\
    
    +GKP + GSA & 71.8 & 42.1 & 9.7 & 42.8 \\
    BCL~\cite{zhu2022balanced} & 67.2 & 53.1 & 32.9 & 51.9 \\
    +GKP & 67.4 & 53.8 & 33.2 & 52.4 \\
    +GSA & 67.3 & 54.0 & 34.1 & 52.7 \\
    \hdashline 
   \textbf{Ours} & \textbf{67.3    }\textsubscript{\scriptsize\textcolor{blue}{$\uparrow$0.1}} & \textbf{54.9}\textsubscript{\scriptsize\textcolor{blue}{$\uparrow$1.8}} & \textbf{34.9}\textsubscript{\scriptsize\textcolor{blue}{$\uparrow$2.0}} & \textbf{53.2}\textsubscript{\scriptsize\textcolor{blue}{$\uparrow$1.3}} \\
    \hline\thickhline
    \end{tabular}%} % <-- 已移除
    \label{tab:Ablation_components}
\end{table}
    
\noindent\textbf{Importance of Gradient Decomposition in GSA.} 
To validate the gradient decomposition (Eq.~\ref{eq:gsa_direction}), we compare the performance of different perturbation directions in Table~\ref{tab:Gradient}. Our GSA, using the group-specific gradient surpasses all other methods, achieves an accuracy 53.2\%. This compares favorably to a standard SAM baseline, which does not process the gradient direction and achieves 52.1\%. While using only the projected component $\operatorname{Proj}_{\nabla_\theta \mathcal{L}_{D}(\boldsymbol{\theta})}  \nabla_\theta \mathcal{L}_{\mathcal{D}_g}{(\boldsymbol{\theta})}$ termed \textrm{GSA-proj}  yields obvious performance degradation of 46.4\%. This performance gap demonstrates that the head-dominated global gradient is a harmful perturbation direction in long-tailed learning.

\begin{table}[ht]
    \caption{Importance of Gradient Decomposition in GSA.}
    \centering
    \small % <-- 使用 \small 字体
    % \resizebox{0.48\textwidth}{!}{ % <-- 已移除
    \begin{tabular}{|l|cccc|}
    \hline\thickhline
    \rowcolor{mygray}
    Method & Many$\uparrow$ & Med.$\uparrow$ & Few$\uparrow$ & All$\uparrow$ \\ 
    \hline\hline % <-- 修正了 \hline \hline
     SAM & 66.3 & 53.0 &34.5 & 52.1\\

     GSA-proj & 64.7 & 43.8 & 28.1 & 46.4\\

    \hdashline 
   \textbf{GSA (Ours)} & \textbf{67.3    }\textsubscript{\scriptsize\textcolor{blue}{$\uparrow$1.0}} & \textbf{54.9}\textsubscript{\scriptsize\textcolor{blue}{$\uparrow$1.9}} & \textbf{34.9}\textsubscript{\scriptsize\textcolor{blue}{$\uparrow$0.4}} & \textbf{53.2}\textsubscript{\scriptsize\textcolor{blue}{$\uparrow$1.1}} \\
    \hline\thickhline
    \end{tabular}%} % <-- 已移除
    \label{tab:Gradient}
\end{table}
    
\subsection{Further Analysis from the Gradient Similarity}
To further validate the effectiveness of our method, we analyze the training dynamics from the gradient similarity~\cite{li2024long} perspective. We measure the mean gradient similarity between class and whole gradient in each batch in different epochs, comparing our method against the Cross-Entropy (CE) baseline. The similarity measures the contribution of gradients from different classes to the gradient descent process, and a larger similarity means a larger contribution.
 
\begin{figure}
    \centering
    \includegraphics[width=1\linewidth]{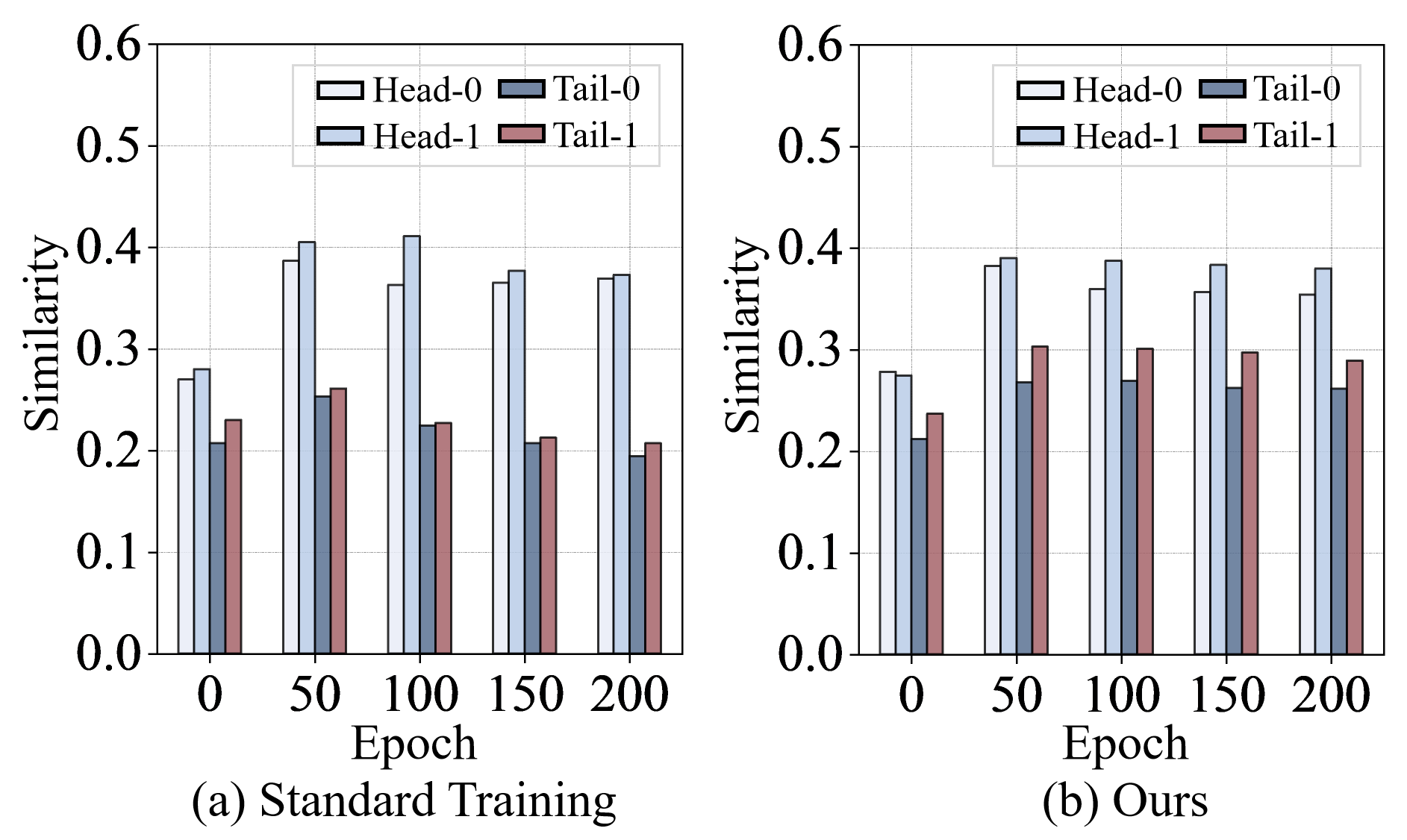}
    \caption{Gradient imbalance in long-tail learning. The bars denote the mean similarity between class-level and batch-level gradients in each batch. This experiment compares our framework against the Cross-Entropy (CE) baseline on the CIFAR10-LT dataset~\cite{cao2019learning}. We show the result of the top two head classes and the last two tail classes.}
    \label{fig:gradient}
\end{figure}

As illustrated in Fig.~\ref{fig:gradient}, while the head classes gradients exhibit high similarity in both methods, their behavior on tail classes diverges significantly. For the baseline method (a), the gradient similarity of the tail classes declines after 50 epochs. In contrast, our method (b) consistently maintains a larger
gradient similarity for the tail classes throughout the training, demonstrating the effectiveness in preserving the tail classes knowledge.

\section{Conclusion}
Long-tail (LT) recognition remained a fundamental challenge in deep learning due to severe class imbalance. To address the ``seesaw dilemma'' in LT, we presented a novel framework from loss landscape view and traced the ``tail performance degradation''. Inspired by continual learning, our method introduced two key innovations: a Grouped Knowledge Preservation module to mitigate ``tail performance degradation'', and a Grouped Sharpness Aware module to seek shared, flat minima. Both modules were enabled by our Memory-based Grouping strategy, which dynamically clusters classes based on their convergence characteristics. Extensive experiments on four benchmark datasets demonstrated our method's superior performance, achieving significant performance gains over state-of-the-art methods without requiring external data. Our proposed method was a general framework for handling ``tail performance degradation'' in data learning, and can be readily extended to other long tail tasks, such as trajectory prediction and object detection.

{
    \small
    \section*{Acknowledgments}

This work was supported by the National Key R\&D Program of China (Program Nos. 2024YFC2707500), the National Natural Science Foundation of China (Grant Nos. T2541059 and 62302046), the Shandong Excellent Young Scientists Fund (ZR2024YQ006), and the Taishan Scholars Program (Grant No. tsqn202507014).
    \bibliographystyle{ieeenat_fullname}
    \bibliography{main}

\begin{thebibliography}{76}
\providecommand{\natexlab}[1]{#1}
\providecommand{\url}[1]{\texttt{#1}}
\expandafter\ifx\csname urlstyle\endcsname\relax
  \providecommand{\doi}[1]{doi: #1}\else
  \providecommand{\doi}{doi: \begingroup \urlstyle{rm}\Url}\fi

\bibitem[Ahn et~al.(2023)Ahn, Ko, and Yun]{ahn2023cuda}
Sumyeong Ahn, Jongwoo Ko, and Se-Young Yun.
\newblock {CUDA}: Curriculum of data augmentation for long-tailed recognition.
\newblock In \emph{ICLR}, 2023.

\bibitem[Bayoudh(2024)]{bayoudh2024survey}
Khaled Bayoudh.
\newblock A survey of multimodal hybrid deep learning for computer vision: Architectures, applications, trends, and challenges.
\newblock \emph{Information Fusion}, 105:\penalty0 102217, 2024.

\bibitem[Breakwell(1959)]{OptimizationTrajectories}
John~V. Breakwell.
\newblock The optimization of trajectories.
\newblock \emph{Journal of the Society for Industrial and Applied Mathematics}, 7\penalty0 (2):\penalty0 215--247, 1959.

\bibitem[Cao et~al.(2019)Cao, Wei, Gaidon, Arechiga, and Ma]{cao2019learning}
Kaidi Cao, Colin Wei, Adrien Gaidon, Nikos Arechiga, and Tengyu Ma.
\newblock Learning imbalanced datasets with label-distribution-aware margin loss.
\newblock In \emph{NeurIPS}, 2019.

\bibitem[Cui et~al.(2021)Cui, Zhong, Liu, Yu, and Jia]{cui2021parametric}
Jiequan Cui, Zhisheng Zhong, Shu Liu, Bei Yu, and Jiaya Jia.
\newblock Parametric contrastive learning.
\newblock In \emph{ICCV}, 2021.

\bibitem[Cui et~al.(2019)Cui, Jia, Lin, Song, and Belongie]{cui2019class}
Yin Cui, Menglin Jia, Tsung-Yi Lin, Yang Song, and Serge Belongie.
\newblock Class-balanced loss based on effective number of samples.
\newblock In \emph{CVPR}, 2019.

\bibitem[Dong et~al.(2022)Dong, Zhou, Yan, and Zuo]{donglpt}
Bowen Dong, Pan Zhou, Shuicheng Yan, and Wangmeng Zuo.
\newblock Lpt: Long-tailed prompt tuning for image classification.
\newblock In \emph{ICLR}, 2022.

\bibitem[Dong et~al.(2017)Dong, Gong, and Zhu]{dong2017class}
Qi Dong, Shaogang Gong, and Xiatian Zhu.
\newblock Class rectification hard mining for imbalanced deep learning.
\newblock In \emph{ICCV}, 2017.

\bibitem[Esteva et~al.(2021)Esteva, Chou, Yeung, Naik, Madani, Mottaghi, Liu, Topol, Dean, and Socher]{esteva2021deep}
Andre Esteva, Katherine Chou, Serena Yeung, Nikhil Naik, Ali Madani, Ali Mottaghi, Yun Liu, Eric Topol, Jeff Dean, and Richard Socher.
\newblock Deep learning-enabled medical computer vision.
\newblock \emph{NPJ digital medicine}, 4\penalty0 (1):\penalty0 5, 2021.

\bibitem[Foret et~al.(2020)Foret, Kleiner, Mobahi, and Neyshabur]{foret2020sharpness}
Pierre Foret, Ariel Kleiner, Hossein Mobahi, and Behnam Neyshabur.
\newblock Sharpness-aware minimization for efficiently improving generalization.
\newblock In \emph{ICLR}, 2020.

\bibitem[Germain et~al.(2016)Germain, Bach, Lacoste, and Lacoste-Julien]{germain2016pac}
Pascal Germain, Francis Bach, Alexandre Lacoste, and Simon Lacoste-Julien.
\newblock Pac-bayesian theory meets bayesian inference.
\newblock In \emph{NeurIPS}, 2016.

\bibitem[He(2024)]{he2024gradient}
Jiangpeng He.
\newblock Gradient reweighting: Towards imbalanced class-incremental learning.
\newblock In \emph{CVPR}, 2024.

\bibitem[He et~al.(2016)He, Zhang, Ren, and Sun]{he2016deep}
Kaiming He, Xiangyu Zhang, Shaoqing Ren, and Jian Sun.
\newblock Deep residual learning for image recognition.
\newblock In \emph{CVPR}, 2016.

\bibitem[He et~al.(2021)He, Wu, and Wei]{he2021distilling}
Yin-Yin He, Jianxin Wu, and Xiu-Shen Wei.
\newblock Distilling virtual examples for long-tailed recognition.
\newblock In \emph{ICCV}, 2021.

\bibitem[Hong et~al.(2021)Hong, Han, Choi, Seo, Kim, and Chang]{hong2021disentangling}
Youngkyu Hong, Seungju Han, Kwanghee Choi, Seokjun Seo, Beomsu Kim, and Buru Chang.
\newblock Disentangling label distribution for long-tailed visual recognition.
\newblock In \emph{CVPR}, 2021.

\bibitem[Huang et~al.(2016)Huang, Li, Loy, and Tang]{huang2016learning}
Chen Huang, Yining Li, Chen~Change Loy, and Xiaoou Tang.
\newblock Learning deep representation for imbalanced classification.
\newblock In \emph{CVPR}, 2016.

\bibitem[Jian et~al.(2025)Jian, Chen, Wang, Yao, Wang, and Wu]{Jian_2025_ICCV}
Zhongquan Jian, Yanhao Chen, Yancheng Wang, Junfeng Yao, Meihong Wang, and Qingqiang Wu.
\newblock Supervised exploratory learning for long-tailed visual recognition.
\newblock In \emph{ICCV}, 2025.

\bibitem[Jiang et~al.(2019)Jiang, Neyshabur, Mobahi, Krishnan, and Bengio]{jiangfantastic}
Yiding Jiang, Behnam Neyshabur, Hossein Mobahi, Dilip Krishnan, and Samy Bengio.
\newblock Fantastic generalization measures and where to find them.
\newblock In \emph{ICLR}, 2019.

\bibitem[Kang et~al.(2020)Kang, Xie, Rohrbach, Yan, Gordo, Feng, and Kalantidis]{kang2019decoupling}
Bingyi Kang, Saining Xie, Marcus Rohrbach, Zhicheng Yan, Albert Gordo, Jiashi Feng, and Yannis Kalantidis.
\newblock Decoupling representation and classifier for long-tailed recognition.
\newblock In \emph{ICLR}, 2020.

\bibitem[Kaushal et~al.(2024)Kaushal, Tammineni, Rana, Sharma, Sridhar, and Chen]{kaushal2024computer}
Sushant Kaushal, Dushyanth~Kumar Tammineni, Priya Rana, Minaxi Sharma, Kandi Sridhar, and Ho-Hsien Chen.
\newblock Computer vision and deep learning-based approaches for detection of food nutrients/nutrition: New insights and advances.
\newblock \emph{Trends in Food Science \& Technology}, 146:\penalty0 104408, 2024.

\bibitem[Keskar et~al.(2017)Keskar, Mudigere, Nocedal, Smelyanskiy, and Tang]{keskar2017large}
Nitish~Shirish Keskar, Dheevatsa Mudigere, Jorge Nocedal, Mikhail Smelyanskiy, and Ping Tak~Peter Tang.
\newblock On large-batch training for deep learning: Generalization gap and sharp minima.
\newblock In \emph{ICLR}, 2017.

\bibitem[Kirkpatrick et~al.(2017)Kirkpatrick, Pascanu, Rabinowitz, Veness, Desjardins, Rusu, Milan, Quan, Ramalho, Grabska-Barwinska, et~al.]{kirkpatrick2017overcoming}
James Kirkpatrick, Razvan Pascanu, Neil Rabinowitz, Joel Veness, Guillaume Desjardins, Andrei~A Rusu, Kieran Milan, John Quan, Tiago Ramalho, Agnieszka Grabska-Barwinska, et~al.
\newblock Overcoming catastrophic forgetting in neural networks.
\newblock \emph{Proceedings of the National Academy of Sciences}, 114\penalty0 (13):\penalty0 3521--3526, 2017.

\bibitem[Li et~al.(2018)Li, Xu, Taylor, Studer, and Goldstein]{li2018visualizing}
Hao Li, Zheng Xu, Gavin Taylor, Christoph Studer, and Tom Goldstein.
\newblock Visualizing the loss landscape of neural nets.
\newblock In \emph{NeurIPS}, 2018.

\bibitem[Li et~al.(2024{\natexlab{a}})Li, Zhikai, Lu, Lan, Cheung, and Huang]{li2024feature}
Mengke Li, HU Zhikai, Yang Lu, Weichao Lan, Yiu-ming Cheung, and Hui Huang.
\newblock Feature fusion from head to tail for long-tailed visual recognition.
\newblock In \emph{AAAI}, 2024{\natexlab{a}}.

\bibitem[Li et~al.(2021)Li, Gong, Liu, Wang, Qiao, and Cheng]{li2021metasaug}
Shuang Li, Kaixiong Gong, Chi~Harold Liu, Yulin Wang, Feng Qiao, and Xinjing Cheng.
\newblock Metasaug: Meta semantic augmentation for long-tailed visual recognition.
\newblock In \emph{CVPR}, 2021.

\bibitem[Li et~al.(2025{\natexlab{a}})Li, Xu, Yang, Wang, Zhang, Cao, and Huang]{li2025focal}
Sicong Li, Qianqian Xu, Zhiyong Yang, Zitai Wang, Linchao Zhang, Xiaochun Cao, and Qingming Huang.
\newblock Focal-sam: Focal sharpness-aware minimization for long-tailed classification.
\newblock In \emph{NeurIPS}, 2025{\natexlab{a}}.

\bibitem[Li et~al.(2024{\natexlab{b}})Li, Zhou, He, Cheng, and Huang]{li2024friendly}
Tao Li, Pan Zhou, Zhengbao He, Xinwen Cheng, and Xiaolin Huang.
\newblock Friendly sharpness-aware minimization.
\newblock In \emph{CVPR}, 2024{\natexlab{b}}.

\bibitem[Li et~al.(2024{\natexlab{c}})Li, Lyu, Shang, Wan, and Feng]{li2024long}
Weiqi Li, Fan Lyu, Fanhua Shang, Liang Wan, and Wei Feng.
\newblock Long-tailed learning as multi-objective optimization.
\newblock In \emph{AAAI}, 2024{\natexlab{c}}.

\bibitem[Li et~al.(2025{\natexlab{b}})Li, Li, and Jia]{Li_2025_ICCV}
Yuhang Li, Zhuying Li, and Yuheng Jia.
\newblock Boosting class representation via semantically related instances for robust long-tailed learning with noisy labels.
\newblock In \emph{ICCV}, 2025{\natexlab{b}}.

\bibitem[Li and Jia(2025)]{li2025conmix}
Zhixin Li and Yuheng Jia.
\newblock Conmix: Contrastive mixup at representation level for long-tailed deep clustering.
\newblock In \emph{ICLR}, 2025.

\bibitem[Lin et~al.(2014)Lin, Maire, Belongie, Hays, Perona, Ramanan, Doll{\'a}r, and Zitnick]{lin2014microsoft}
Tsung-Yi Lin, Michael Maire, Serge Belongie, James Hays, Pietro Perona, Deva Ramanan, Piotr Doll{\'a}r, and C~Lawrence Zitnick.
\newblock Microsoft coco: Common objects in context.
\newblock In \emph{ECCV}, 2014.

\bibitem[Lin et~al.(2017)Lin, Goyal, Girshick, He, and Doll{\'a}r]{lin2017focal}
Tsung-Yi Lin, Priya Goyal, Ross Girshick, Kaiming He, and Piotr Doll{\'a}r.
\newblock Focal loss for dense object detection.
\newblock In \emph{ICCV}, 2017.

\bibitem[Liu et~al.(2021)Liu, Li, Kang, Hua, and Vasconcelos]{liu2021gistnet}
Bo Liu, Haoxiang Li, Hao Kang, Gang Hua, and Nuno Vasconcelos.
\newblock Gistnet: a geometric structure transfer network for long-tailed recognition.
\newblock In \emph{ICCV}, 2021.

\bibitem[Liu et~al.(2008)Liu, Wu, and Zhou]{liu2008exploratory}
Xu-Ying Liu, Jianxin Wu, and Zhi-Hua Zhou.
\newblock Exploratory undersampling for class-imbalance learning.
\newblock \emph{IEEE Transactions on Systems, Man, and Cybernetics, Part B (Cybernetics)}, 39\penalty0 (2):\penalty0 539--550, 2008.

\bibitem[Liu et~al.(2022)Liu, Mai, Chen, Hsieh, and You]{liu2022towards}
Yong Liu, Siqi Mai, Xiangning Chen, Cho-Jui Hsieh, and Yang You.
\newblock Towards efficient and scalable sharpness-aware minimization.
\newblock In \emph{CVPR}, 2022.

\bibitem[Liu et~al.(2025)Liu, Yang, and Wang]{Liu_2025_ICCV}
Yuting Liu, Liu Yang, and Yu Wang.
\newblock Long-tailed classification with multi-granularity semantics.
\newblock In \emph{ICCV}, 2025.

\bibitem[Liu et~al.(2019)Liu, Miao, Zhan, Wang, Gong, and Yu]{liu2019large}
Ziwei Liu, Zhongqi Miao, Xiaohang Zhan, Jiayun Wang, Boqing Gong, and Stella~X Yu.
\newblock Large-scale long-tailed recognition in an open world.
\newblock In \emph{CVPR}, 2019.

\bibitem[Majee et~al.(2023)Majee, Kothawade, Killamsetty, and Iyer]{majeescore}
Anay Majee, Suraj~Nandkishor Kothawade, Krishnateja Killamsetty, and Rishabh~K Iyer.
\newblock Score: Submodular combinatorial representation learning.
\newblock In \emph{ICML}, 2023.

\bibitem[Narayan et~al.(2025)Narayan, Vs, and Patel]{narayan2025segface}
Kartik Narayan, Vibashan Vs, and Vishal~M Patel.
\newblock Segface: Face segmentation of long-tail classes.
\newblock In \emph{AAAI}, 2025.

\bibitem[Ng et~al.(2001)Ng, Jordan, and Weiss]{ng2001spectral}
Andrew Ng, Michael Jordan, and Yair Weiss.
\newblock On spectral clustering: Analysis and an algorithm.
\newblock In \emph{NeurIPS}, 2001.

\bibitem[Ramanathan et~al.(2020)Ramanathan, Wang, and Mahajan]{ramanathan2020dlwl}
Vignesh Ramanathan, Rui Wang, and Dhruv Mahajan.
\newblock Dlwl: Improving detection for lowshot classes with weakly labelled data.
\newblock In \emph{CVPR}, 2020.

\bibitem[Rangwani et~al.(2022)Rangwani, Aithal, Mishra, et~al.]{rangwani2022escaping}
Harsh Rangwani, Sumukh~K Aithal, Mayank Mishra, et~al.
\newblock Escaping saddle points for effective generalization on class-imbalanced data.
\newblock In \emph{NeurIPS}, 2022.

\bibitem[Reeb et~al.(2018)Reeb, Doerr, Gerwinn, and Rakitsch]{reeb2018learning}
David Reeb, Andreas Doerr, Sebastian Gerwinn, and Barbara Rakitsch.
\newblock Learning gaussian processes by minimizing pac-bayesian generalization bounds.
\newblock In \emph{NeurIPS}, 2018.

\bibitem[Ren et~al.(2020)Ren, Yu, Ma, Zhao, Yi, et~al.]{ren2020balanced}
Jiawei Ren, Cunjun Yu, Xiao Ma, Haiyu Zhao, Shuai Yi, et~al.
\newblock Balanced meta-softmax for long-tailed visual recognition.
\newblock In \emph{NeurIPS}, 2020.

\bibitem[Samuel and Chechik(2021)]{samuel2021distributional}
Dvir Samuel and Gal Chechik.
\newblock Distributional robustness loss for long-tail learning.
\newblock In \emph{ICCV}, 2021.

\bibitem[Shao et~al.(2024)Shao, Zhu, Zhang, and Wu]{shao2024diffult}
Jie Shao, Ke Zhu, Hanxiao Zhang, and Jianxin Wu.
\newblock Diffult: Diffusion for long-tail recognition without external knowledge.
\newblock In \emph{NeurIPS}, 2024.

\bibitem[Shi and Malik(2000)]{shi2000normalized}
Jianbo Shi and Jitendra Malik.
\newblock Normalized cuts and image segmentation.
\newblock \emph{IEEE TPAMI}, 22\penalty0 (8):\penalty0 888--905, 2000.

\bibitem[Shi et~al.(2023)Shi, Wei, Zhou, Han, Shao, and Li]{shi2023parameter}
Jiang-Xin Shi, Tong Wei, Zhi Zhou, Xin-Yan Han, Jie-Jing Shao, and Yufeng Li.
\newblock Parameter-efficient long-tailed recognition.
\newblock \emph{CoRR}, 2023.

\bibitem[Smith et~al.(2024)Smith, Valkov, Halbe, Gutta, Feris, Kira, and Karlinsky]{Smith_2024_CVPR}
James~Seale Smith, Lazar Valkov, Shaunak Halbe, Vyshnavi Gutta, Rogerio Feris, Zsolt Kira, and Leonid Karlinsky.
\newblock Adaptive memory replay for continual learning.
\newblock In \emph{CVPR}, 2024.

\bibitem[Song et~al.(2025)Song, Qu, Zhou, and Cheng]{Song_2025_CVPR}
Mingyang Song, Xiaoye Qu, Jiawei Zhou, and Yu Cheng.
\newblock From head to tail: Towards balanced representation in large vision-language models through adaptive data calibration.
\newblock In \emph{CVPR}, 2025.

\bibitem[Tang et~al.(2020)Tang, Huang, and Zhang]{tang2020long}
Kaihua Tang, Jianqiang Huang, and Hanwang Zhang.
\newblock Long-tailed classification by keeping the good and removing the bad momentum causal effect.
\newblock In \emph{NeurIPS}, 2020.

\bibitem[Van~de Ven and Tolias(2019)]{van2019three}
Gido~M Van~de Ven and Andreas~S Tolias.
\newblock Three scenarios for continual learning.
\newblock \emph{arXiv preprint arXiv:1904.07734}, 2019.

\bibitem[Van~Horn et~al.(2018)Van~Horn, Mac~Aodha, Song, Cui, Sun, Shepard, Adam, Perona, and Belongie]{van2018inaturalist}
Grant Van~Horn, Oisin Mac~Aodha, Yang Song, Yin Cui, Chen Sun, Alex Shepard, Hartwig Adam, Pietro Perona, and Serge Belongie.
\newblock The inaturalist species classification and detection dataset.
\newblock In \emph{CVPR}, 2018.

\bibitem[Wang et~al.(2024{\natexlab{a}})Wang, Zhang, Su, and Zhu]{wang2024comprehensive}
Liyuan Wang, Xingxing Zhang, Hang Su, and Jun Zhu.
\newblock A comprehensive survey of continual learning: Theory, method and application.
\newblock \emph{IEEE TPAMI}, 46\penalty0 (8):\penalty0 5362--5383, 2024{\natexlab{a}}.

\bibitem[Wang et~al.(2024{\natexlab{b}})Wang, Zhao, Wen, Wang, Wang, Zhang, and Wang]{DBLP:conf/nips/Wang0WWW0024}
Pengkun Wang, Zhe Zhao, Haibin Wen, Fanfu Wang, Binwu Wang, Qingfu Zhang, and Yang Wang.
\newblock Llm-autoda: Large language model-driven automatic data augmentation for long-tailed problems.
\newblock In \emph{NeurIPS}, 2024{\natexlab{b}}.

\bibitem[Wang et~al.(2021)Wang, Lian, Miao, Liu, and Yu]{wang2020long}
Xudong Wang, Long Lian, Zhongqi Miao, Ziwei Liu, and Stella Yu.
\newblock Long-tailed recognition by routing diverse distribution-aware experts.
\newblock In \emph{ICLR}, 2021.

\bibitem[Wang et~al.(2017)Wang, Ramanan, and Hebert]{wang2017learning}
Yu-Xiong Wang, Deva Ramanan, and Martial Hebert.
\newblock Learning to model the tail.
\newblock In \emph{NeurIPS}, 2017.

\bibitem[Wang et~al.(2023)Wang, Xu, Yang, He, Cao, and Huang]{wang2023unified}
Zitai Wang, Qianqian Xu, Zhiyong Yang, Yuan He, Xiaochun Cao, and Qingming Huang.
\newblock A unified generalization analysis of re-weighting and logit-adjustment for imbalanced learning.
\newblock In \emph{NeurIPS}, 2023.

\bibitem[Wei et~al.(2021)Wei, Sohn, Mellina, Yuille, and Yang]{wei2021crest}
Chen Wei, Kihyuk Sohn, Clayton Mellina, Alan Yuille, and Fan Yang.
\newblock Crest: A class-rebalancing self-training framework for imbalanced semi-supervised learning.
\newblock In \emph{CVPR}, 2021.

\bibitem[Wu et~al.(2020)Wu, Morgado, Wang, Ho, and Vasconcelos]{wu2020solving}
Tz-Ying Wu, Pedro Morgado, Pei Wang, Chih-Hui Ho, and Nuno Vasconcelos.
\newblock Solving long-tailed recognition with deep realistic taxonomic classifier.
\newblock In \emph{ECCV}, 2020.

\bibitem[Xie et~al.(2017)Xie, Girshick, Doll{\'a}r, Tu, and He]{xie2017aggregated}
Saining Xie, Ross Girshick, Piotr Doll{\'a}r, Zhuowen Tu, and Kaiming He.
\newblock Aggregated residual transformations for deep neural networks.
\newblock In \emph{CVPR}, 2017.

\bibitem[Yang et~al.(2022)Yang, Jiang, Song, and Guo]{yang2022survey}
Lu Yang, He Jiang, Qing Song, and Jun Guo.
\newblock A survey on long-tailed visual recognition.
\newblock \emph{IJCV}, 130\penalty0 (7):\penalty0 1837--1872, 2022.

\bibitem[Yi et~al.(2025)Yi, Yao, Lyu, Ling, Douady, and Chen]{yi2025geometry}
Lingjie Yi, Jiachen Yao, Weimin Lyu, Haibin Ling, Raphael Douady, and Chao Chen.
\newblock Geometry of long-tailed representation learning: Rebalancing features for skewed distributions.
\newblock In \emph{ICLR}, 2025.

\bibitem[Zhang et~al.(2021{\natexlab{a}})Zhang, Pan, Li, Hu, Xuan, Changpinyo, Gong, and Chao]{zhang2021mosaicos}
Cheng Zhang, Tai-Yu Pan, Yandong Li, Hexiang Hu, Dong Xuan, Soravit Changpinyo, Boqing Gong, and Wei-Lun Chao.
\newblock Mosaicos: A simple and effective use of object-centric images for long-tailed object detection.
\newblock In \emph{ICCV}, 2021{\natexlab{a}}.

\bibitem[Zhang et~al.(2021{\natexlab{b}})Zhang, Li, Yan, He, and Sun]{zhang2021distribution}
Songyang Zhang, Zeming Li, Shipeng Yan, Xuming He, and Jian Sun.
\newblock Distribution alignment: A unified framework for long-tail visual recognition.
\newblock In \emph{CVPR}, 2021{\natexlab{b}}.

\bibitem[Zhang et~al.(2021{\natexlab{c}})Zhang, Wei, Zhou, and Wu]{zhang2021bag}
Yongshun Zhang, Xiu-Shen Wei, Boyan Zhou, and Jianxin Wu.
\newblock Bag of tricks for long-tailed visual recognition with deep convolutional neural networks.
\newblock In \emph{AAAI}, 2021{\natexlab{c}}.

\bibitem[Zhang et~al.(2023{\natexlab{a}})Zhang, Kang, Hooi, Yan, and Feng]{zhang2023deep}
Yifan Zhang, Bingyi Kang, Bryan Hooi, Shuicheng Yan, and Jiashi Feng.
\newblock Deep long-tailed learning: A survey.
\newblock \emph{IEEE TPAMI}, 45\penalty0 (9):\penalty0 10795--10816, 2023{\natexlab{a}}.

\bibitem[Zhang et~al.(2023{\natexlab{b}})Zhang, Zhou, Hooi, Wang, and Feng]{zhang2023expanding}
Yifan Zhang, Daquan Zhou, Bryan Hooi, Kai Wang, and Jiashi Feng.
\newblock Expanding small-scale datasets with guided imagination.
\newblock In \emph{NeurIPS}, 2023{\natexlab{b}}.

\bibitem[Zhang and Pfister(2021)]{zhang2021learning}
Zizhao Zhang and Tomas Pfister.
\newblock Learning fast sample re-weighting without reward data.
\newblock In \emph{ICCV}, 2021.

\bibitem[Zhao et~al.(2024)Zhao, Dai, Li, Hu, Zhang, and Liu]{zhao2024ltgc}
Qihao Zhao, Yalun Dai, Hao Li, Wei Hu, Fan Zhang, and Jun Liu.
\newblock Ltgc: Long-tail recognition via leveraging llms-driven generated content.
\newblock In \emph{CVPR}, 2024.

\bibitem[Zhou et~al.(2020)Zhou, Cui, Wei, and Chen]{zhou2020bbn}
Boyan Zhou, Quan Cui, Xiu-Shen Wei, and Zhao-Min Chen.
\newblock Bbn: Bilateral-branch network with cumulative learning for long-tailed visual recognition.
\newblock In \emph{CVPR}, 2020.

\bibitem[Zhou et~al.(2025)Zhou, Wu, and Yang]{zhou2025class}
Xiaoling Zhou, Ou Wu, and Nan Yang.
\newblock Class and attribute-aware logit adjustment for generalized long-tail learning.
\newblock In \emph{AAAI}, 2025.

\bibitem[Zhou et~al.(2023{\natexlab{a}})Zhou, Qu, Xu, and Shen]{zhou2023imbsam}
Yixuan Zhou, Yi Qu, Xing Xu, and Hengtao Shen.
\newblock Imbsam: A closer look at sharpness-aware minimization in class-imbalanced recognition.
\newblock In \emph{ICCV}, 2023{\natexlab{a}}.

\bibitem[Zhou et~al.(2023{\natexlab{b}})Zhou, Li, Zhao, Heng, and Gong]{zhou2023class}
Zhipeng Zhou, Lanqing Li, Peilin Zhao, Pheng-Ann Heng, and Wei Gong.
\newblock Class-conditional sharpness-aware minimization for deep long-tailed recognition.
\newblock In \emph{CVPR}, 2023{\natexlab{b}}.

\bibitem[Zhu et~al.(2022)Zhu, Wang, Chen, Chen, and Jiang]{zhu2022balanced}
Jianggang Zhu, Zheng Wang, Jingjing Chen, Yi-Ping~Phoebe Chen, and Yu-Gang Jiang.
\newblock Balanced contrastive learning for long-tailed visual recognition.
\newblock In \emph{CVPR}, 2022.

\bibitem[Zhu et~al.(2024)Zhu, Fu, Shao, Liu, and Wu]{zhu2024rectify}
Ke Zhu, Minghao Fu, Jie Shao, Tianyu Liu, and Jianxin Wu.
\newblock Rectify the regression bias in long-tailed object detection.
\newblock In \emph{ECCV}, 2024.

\end{thebibliography}
}

% % WARNING: do not forget to delete the supplementary pages from your submission 
\clearpage
\setcounter{page}{1}
\maketitlesupplementary

\section{More Experiment Protocols}

\subsection{Implementation Details}
\label{supp:Details}
Our implementation follows~\cite{zhu2022balanced}. For both CIFAR-10-LT and CIFAR-100-LT, we adopt ResNet-32 as the backbone. 
Our model is trained for 200 epochs with a batch size of 256 based on an SGD optimizer. 
The momentum is set to 0.9 and the weight decay is $5 \times 10^{-4}$. 
The learning rate warms up to 0.15 in the first 5 epochs and decays by 0.1 at the 160 and 180 epochs.

We adopt the ResNet-50~\cite{he2016deep} architecture as the model backbone for both 
ImageNet-LT and iNaturalist 2018. The model is optimized based on SGD with a 
fixed momentum of 0.9 and a batch size of 256. For ImageNet-LT, we train the model 
for 90 epochs using an initial learning rate of 0.1 and a weight decay of $5 \times 10^{-4}$. 
For iNaturalist 2018, the training is extended to 100 epochs, with an initial learning 
rate of 0.2 and a weight decay of $1 \times 10^{-4}$. A cosine scheduler is employed for 
learning rate adjustments across all experiments. 
% --- 修改后的 Subsection 1 ---

\subsection{Definition of Feature Quality}
\label{supp:Q_Detail}

As stated in the main paper (Eq.1), the memory bank $\mathcal{M}$ is populated by storing the encoder parameters $\theta_{enc}^c$ that yield the highest feature quality $Q$. We provide the mathematical formulation for $Q$, which is based on the SCoRe framework~\cite{majeescore}.

First, the inter-class separation for class $c$ is computed as the minimum distance to any other class centroid:
\begin{equation}
\text{Dis}(\theta_{enc}, c) = \min_{c' \neq c} ||\mu_c - \mu_{c'}||_2,
\end{equation}
Second, the intra-class variance is computed as:
\begin{equation}
\text{Var}(\theta_{enc}, c) = \frac{1}{|\mathcal{A}_c|} \sum_{a \in \mathcal{A}_c} ||a - \mu_c||_2^2,
\end{equation}
where $\mathcal{A}_c$ is the set of feature vectors $a$ (where $a = f(\theta_{enc}, x)$, Sec.~\ref{sec:over}) for class $c$, and $\mu_c$ is the feature centroid. $\mathcal{A}_c$ and $\mu_c$ are computed using the encoder state $\theta_{enc}$.

The final quality score $Q(\theta_{enc}, c)$ is then calculated as:
\begin{equation}
Q(\theta_{enc}, c) = \text{Dis}(\theta_{enc}, c) - \beta \log(\text{Var}(\theta_{enc}, c)),
\end{equation}
where $\beta$ is a hyperparameter balancing the two components, which is set to 0.5 in our experiments.

\subsection{Definition of Adaptive parameter}
The adaptive parameter scheduled  $\alpha$ (Eq.~\ref{eq:Loss}) at epoch $t$ is updated according to the cosine annealing schedule:
\begin{equation}
    \label{eq:alpha_schedule}
    \alpha = \alpha_{end} + \frac{1}{2}(\alpha_{start} - \alpha_{end}) \left( 1 + \cos\left( \frac{t \pi}{T} \right) \right),
\end{equation}
where $\alpha_{start} = 0.95$ and $\alpha_{end} = 0.6$ denote the initial and final values respectively, and $T$ represents the total number of training epochs.

\section{More Experiment Results}
\subsection{Ablation Study About Group Numbers}
\begin{figure}
    \centering
    \includegraphics[width=1\linewidth]{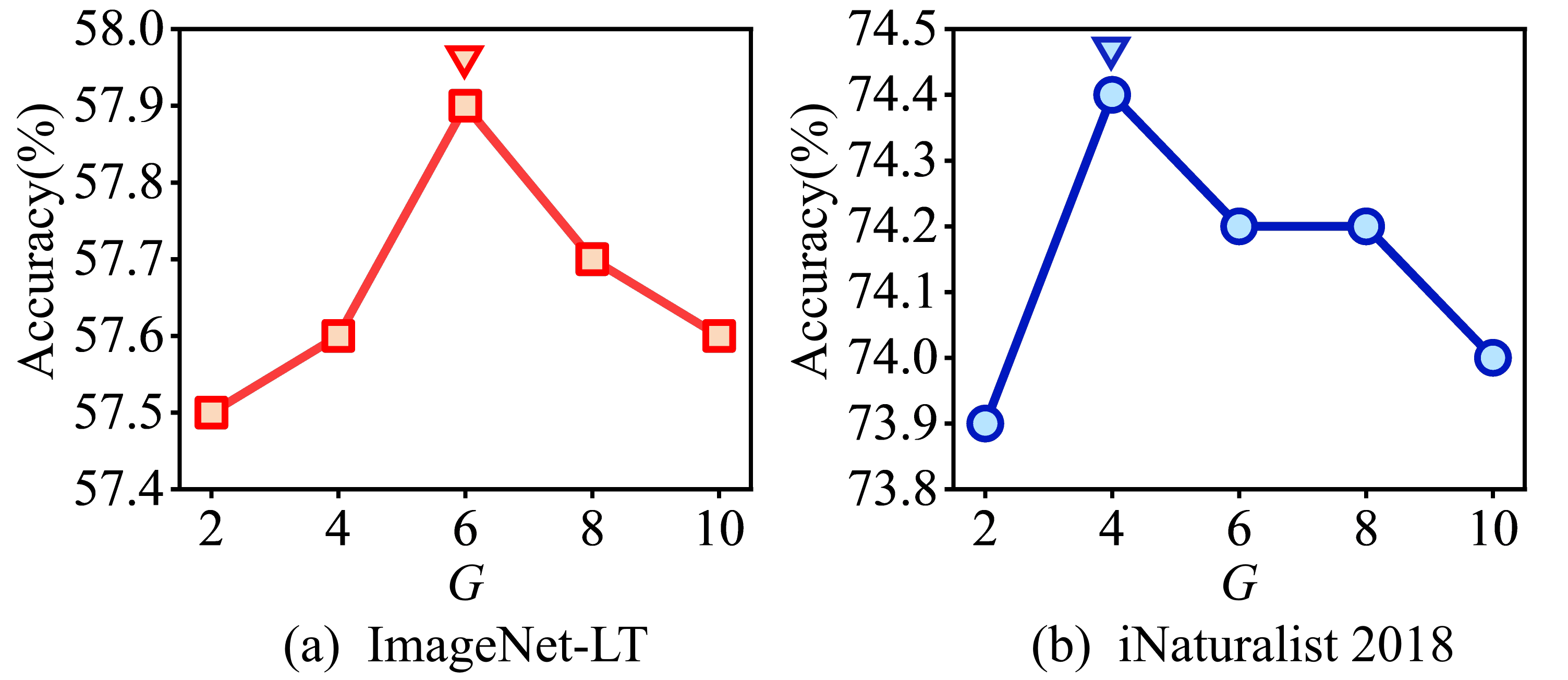}
    \caption{Ablation study about group numbers on ImageNet-LT and iNaturalist 2018.}
    \label{fig:groupNum}
\end{figure}

    % grouping in imageNetLt and inaul ablation
We conduct a detailed ablation study on the impact of the group number $G$ on the large-scale datasets, ImageNet-LT and iNaturalist 2018.We report the Top-1 accuracy for various $G$ values in Fig.~\ref{fig:groupNum}. The performance peaks at $G=4$ for iNaturalist 2018, whereas ImageNet-LT achieves its best results at $G=6$. Similar to our other experiments, employing an excessive number of groups does not yield further gains.

\begin{table}[ht]
  \centering
  \caption{Hyperparameter ablation analysis for the knowledge preservation strength $\lambda$ on CIFAR100-LT ($r=100$).}
  % 标签
  \label{tab:lambda_sensitivity}
  \begin{tabular}{l c}
    \toprule
    \textbf{Hyperparameter ($\lambda$)}  & \textbf{Overall Acc.} \\
    \midrule
    10 & 52.6 \\
    50    & 52.7 \\
    \textbf{100} & \textbf{53.2} \\
    500 & 52.4 \\
    1000 & 52.1 \\
    \bottomrule
  \end{tabular}
\end{table}
\subsection{Ablation Study About Knowledge Preservation Strength}
    % Strength of Knowledge Preservation
We analyze the model's sensitivity to the knowledge preservation strength, controlled by the hyperparameter $\lambda$. We tested a range of $\lambda$ values ($\lambda \in \{10, 50, 100, 500, 1000\}$) on CIFAR100-LT ($r=100$).  As shown in Table~\ref{tab:lambda_sensitivity}, a small $\lambda$ (\eg, $0.1$) is insufficient to prevent catastrophic forgetting, resulting in low overall accuracy. Conversely, a large $\lambda$ (\eg, $1000$) severely hinders the acquisition of new knowledge for the current group. Our model achieves the optimal balance at $\lambda = 100$, resulting in an accuracy of 53.2\%.

    \subsection{Ablation Study About Perturbation Scale}
    % perturbation ablation
   To further validate our derived characteristic grouped radius, we empirically study the influence of varying the perturbation scale.

   To investigate the optimal perturbation scale, we multiply Eq.~\ref{eq:group_radius} by a scaling factor $Z=\rho_g/\rho_g^*$, varying the norm from $10^{-1}$ to $10^{-7}$. As shown in Figure 1, the performance of GSA peaks when $Z$ is set to $10^{-2}$, confirming the existence of an optimal scale. Nevertheless, the coefficient for the characteristic grouped radius $\rho_g^*$ (Eq.~\ref{eq:group_radius}) is not a precise value and requires empirical tuning (as further elaborated in the Remarks of Sec.~\ref{Radius}).
   \begin{figure}[ht]
    \centering
    \includegraphics[width=0.8\linewidth]{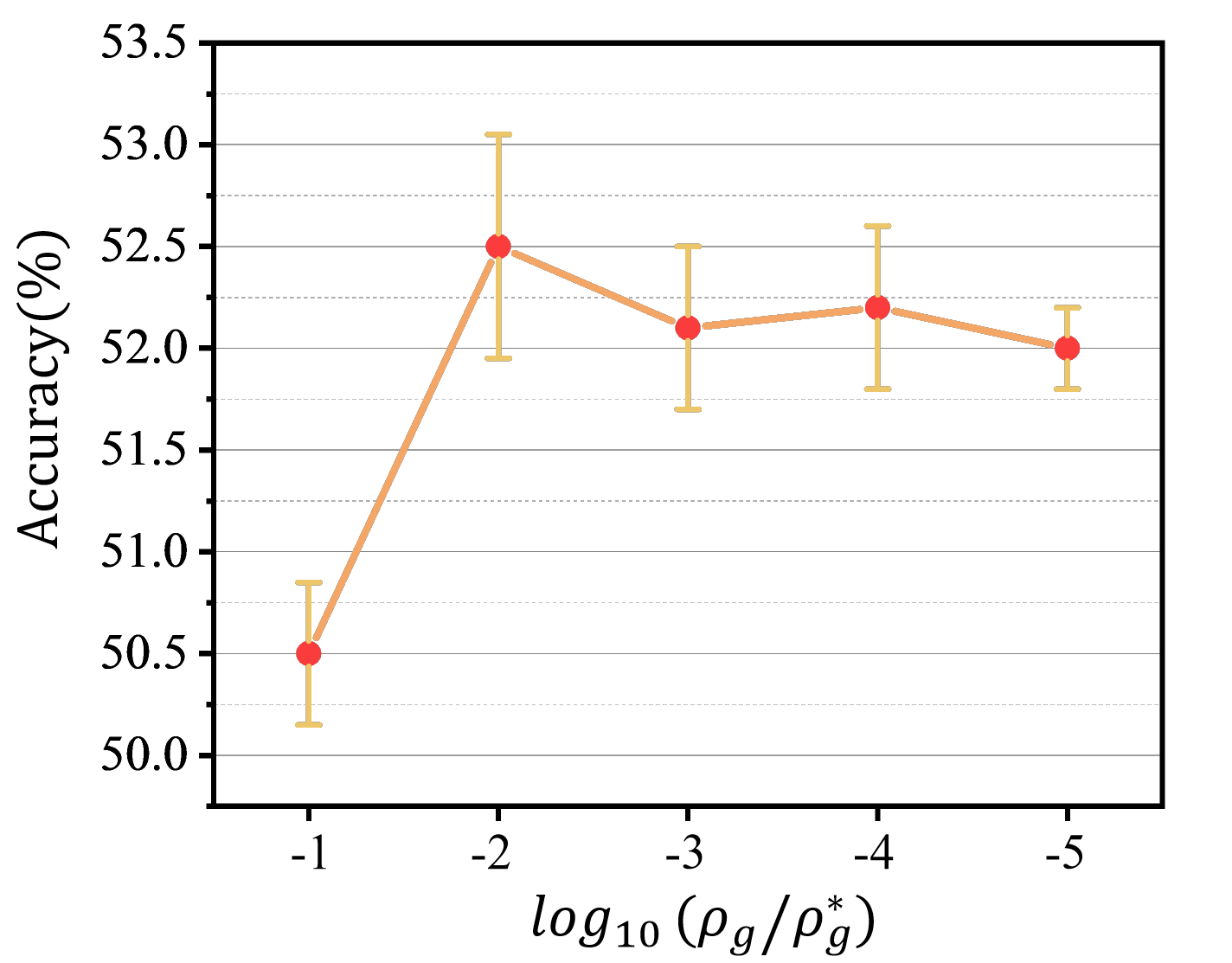}
    \caption{The impact of the perturbation scale.}
    \label{fig:perturbation}
\end{figure}

\begin{figure*}[ht]
    \centering
    % --- 第一行: HEADs (ce_head, sam_head, our_head) ---
    
    % ce_head.png
    \begin{subfigure}[b]{0.32\textwidth}
        \centering
        \includegraphics[width=\textwidth]{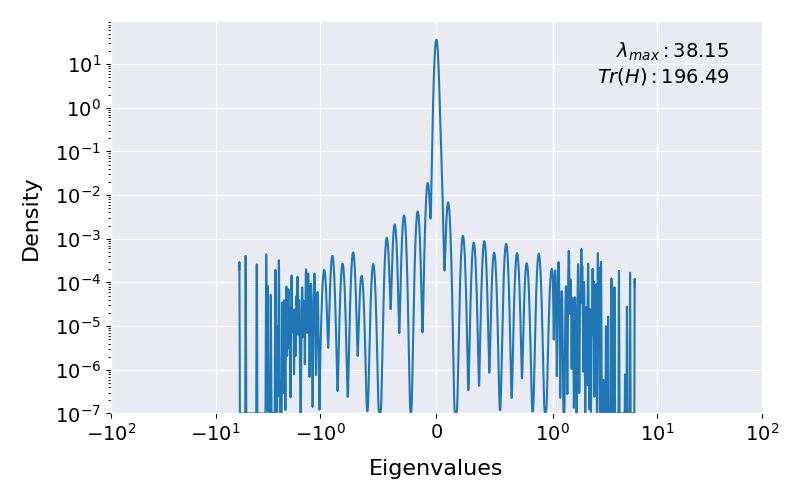}
        \caption{CE: Head Class}
        \label{fig:ce_head}
    \end{subfigure}
    \hfill
    % sam_head.png
    \begin{subfigure}[b]{0.32\textwidth}
        \centering
        \includegraphics[width=\textwidth]{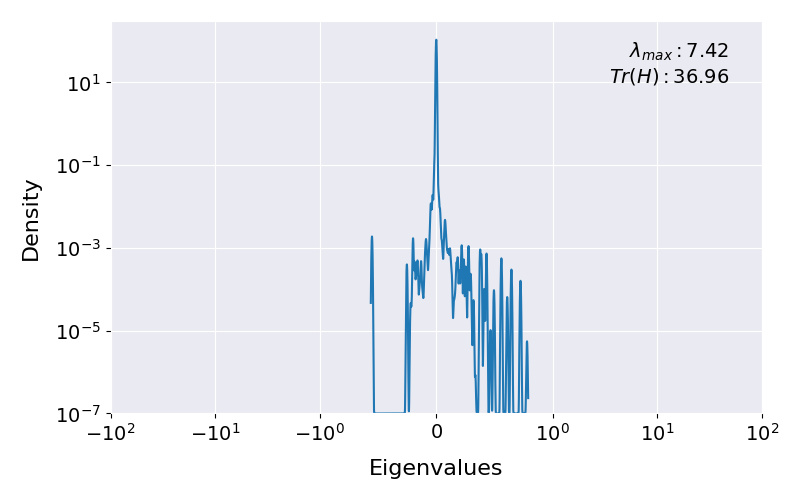}
        \caption{SAM: Head Class}
        \label{fig:sam_head}
    \end{subfigure}
    \hfill
    % our_head.png (formerly ourSam_head.png)
    \begin{subfigure}[b]{0.32\textwidth}
        \centering
        \includegraphics[width=\textwidth]{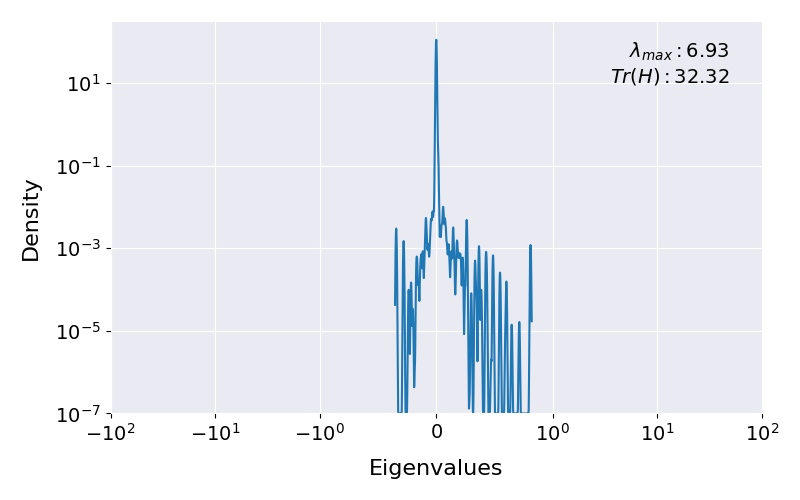} % 假设文件名为 our_head.png
        \caption{Our method: Head Class}
        \label{fig:our_head}
    \end{subfigure}
    
    \vspace{1em} % 增加行间距

    % --- 第二行: TAILs (ce_tail, sam_tail, our_tail) ---
    
    % ce_tail.png
    \begin{subfigure}[b]{0.32\textwidth}
        \centering
        \includegraphics[width=\textwidth]{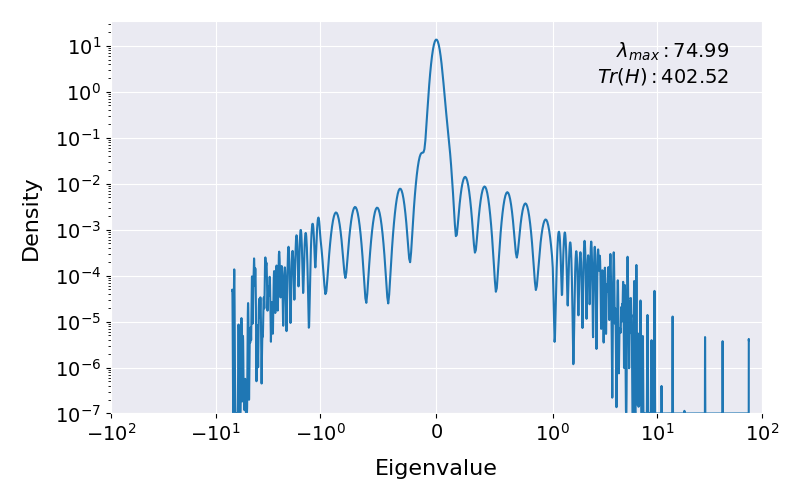}
        \caption{CE: Tail Class}
        \label{fig:ce_tail}
    \end{subfigure}
    \hfill
    % sam_tail.png
    \begin{subfigure}[b]{0.32\textwidth}
        \centering
        \includegraphics[width=\textwidth]{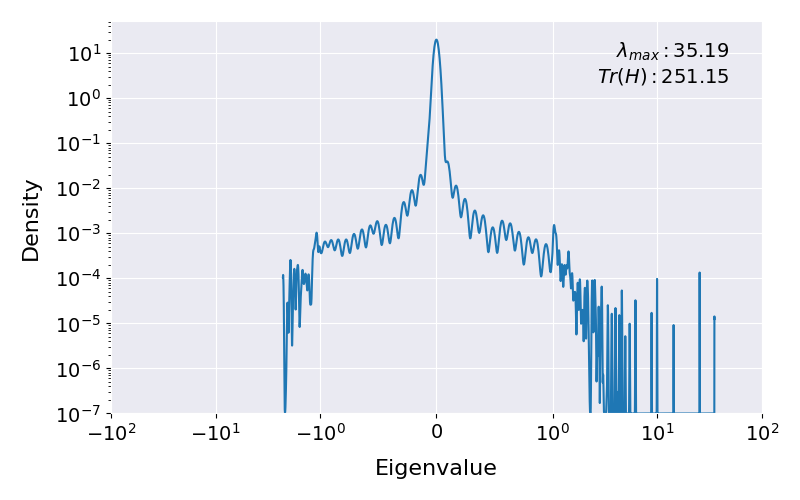}
        \caption{SAM: Tail Class}
        \label{fig:sam_tail}
    \end{subfigure}
    \hfill
    % our_tail.png (formerly ourSam_tail.png)
    \begin{subfigure}[b]{0.32\textwidth}
        \centering
        \includegraphics[width=\textwidth]{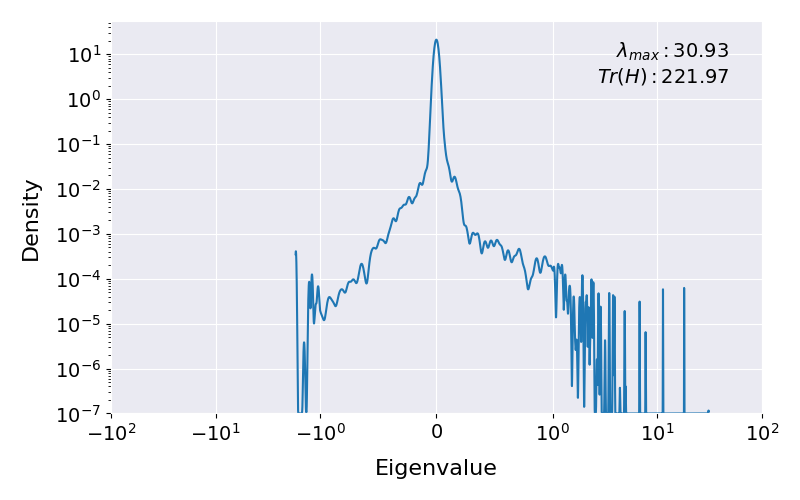} % 假设文件名为 our_tail.png
        \caption{Our method: Tail Class}
        \label{fig:our_tail}
    \end{subfigure}
    
    \caption{Eigenvalue density distributions for head and tail layers, arranged by CE (the baseline method with the naive  Cross Entropy loss), SAM~\cite{foret2020sharpness}, and our method.} % 主图注放在这里
    \label{fig:all_eigenvalues}

\end{figure*}
    \subsection{Analysis About Loss Landscape}
    % from Eigen Spectral Density of Hessian to test GSA
    This section presents additional results of the spectral density of hessian for ResNet models trained with CE (the baseline method with the naive  Cross Entropy loss), CE+SAM~\cite{foret2020sharpness} and CE+Ours. We analyze models trained on CIFAR-10 LT datasets using VS and CE loss
functions. 

Experimental results demonstrate that, compared to SAM, our method achieves lower values for both the largest eigenvalue ($\lambda_{\text{max}}$) and the trace ($\text{tr}({ht})$) of the Hessian matrix across both head and tail classes. This indicates that our approach successfully yields a flatter loss surface.

\subsection{Analysis About Training Time Cost}
    % training Cost
The wall-clock training time of our method ($G=4$) was assessed against the CE and BCL baselines (Table~\ref{tab:time}). Our framework required 1.67 hours to train, resulting in an additional 0.62 hours compared to the BCL baseline.

\begin{table}[ht]
  % \centering 命令使表格内容 (tabular) 在栏宽内居中
  \centering
  
  % 标题
  \caption{Analysis about training time cost.}
  % 标签
  \label{tab:time}
  \begin{tabular}{l cc}
    \toprule
    \textbf{CE}  & \textbf{BCL} & \textbf{Ours} \\
    \midrule
    0.52h & 1.05h & 1.67h\\
 
    \bottomrule
  \end{tabular}
\end{table}
\section{Theoretical Proofs}
\label{supp:Proof}

\subsection{Theoretical Proofs About Convergence}
We present the convergence analysis for our framework (Eq. 16), proving the algorithm converges under standard non-convex stochastic optimization assumptions.

Our proof builds upon the convergence analysis of F-SAM~\cite{li2024friendly}, which established convergence for a min-max sharpness objective. We demonstrate that the inclusion of our Grouped Knowledge Preservation (GKP) module, which acts as a convex regularizer, preserves and stabilizes this convergence.

\subsubsection{Objective Function and Update Rule}
First, We define the complete objective function $\mathcal{L}(\theta)$ as:
\begin{equation}
    \label{supp:objection} % (我为你添加了 'supp:' 前缀以避免冲突)
\mathcal{L}(\theta) = \sum_{g=1}^G \left[ \alpha \left( \mathcal{L}_{gsa}^g(\theta) \right) + (1-\alpha) \mathcal{L}_{gkp}^g(\theta) \right],
\end{equation}
where $G$ is the total number of groups, $\mathcal{L}_{gsa}^g(\theta)$ is the loss for group $g$, and $\mathcal{L}_{gkp}^g(\theta)$ is the GKP loss (Eq.6).

Second, we define our framework gradient $g_t'$. This is the stochastic gradient of our full objective $\mathcal{L}$ (Eq.16), computed on a mini-batch $\mathcal{S}$ at step $t$:
\begin{equation}
    \label{supp:objectionBatch}
g_t' =  \sum_{g=1}^G \left[ \alpha \nabla_{\theta} \mathcal{L}_{\mathcal{S},gsa}^g(\theta_t) + (1-\alpha) \nabla_{\theta} \mathcal{L}_{\mathcal{S},gkp}^g(\theta_t) \right],
\end{equation}
expanding $\mathcal{L}_{\mathcal{S},gsa}^g$ using its definition from Eq.15:
\begin{equation}
\label{supp:objectionBatch}
\begin{split}
    g_t' = \sum_{g=1}^G \bigg[ & \alpha \left( \nabla_{\theta} \mathcal{L}_{\mathcal{S},gsa}^g(\theta_t + \hat{\epsilon}_g^*) + \nabla_{\theta} \mathcal{L}_{\mathcal{S},reg}^g(\theta_t) \right) \\
    & + (1-\alpha) \nabla_{\theta} \mathcal{L}_{\mathcal{S},gkp}^g(\theta_t) \bigg],
\end{split}
\end{equation}
with the update rule is $\theta_{t+1} = \theta_t - \eta_t g_t'$.

\subsubsection{Core Assumptions}
We adopt standard assumptions from stochastic non-convex optimization~\cite{li2024friendly}:
\begin{itemize}
    \item $\beta$-Smoothness: The objective $\mathcal{L}(\theta)$ is assumed to be $\beta$-smooth.
    \item Bounded Gradients: The stochastic gradients of all components are assumed to be bounded by positive constants $M$:
    \begin{itemize}
        \item $\mathbb{E}[\|\nabla \mathcal{L}_{\mathcal{S},gsa}^g(\theta)\|_2^2] \le M_{gsa}^2$
        \item $\|\nabla_{\theta} \mathcal{L}_{\mathcal{S},reg}^g(\theta)\|_2^2 \le M_{reg}^2$
        \item $\mathbb{E}[\|\nabla_{\theta} \mathcal{L}_{\mathcal{S},gkp}^g(\theta)\|_2^2] \le M_{gkp}^2$
    \end{itemize}
    \item Bounded Perturbation: The GSA perturbation vector $\hat{\epsilon}_g^*$ (Eq.14) is bounded by the radius $\rho_t$: $\|\hat{\epsilon}_g^*\| \le \rho_t$ for all $g \in \{1, ..., G\}$.
\end{itemize}

\subsubsection{Convergence Derivation}
We begin with the standard $\beta$-smoothness inequality for the objective $\mathcal{L}(\theta)$, taking expectation $\mathbb{E}_t$ conditioned on $\theta_t$:
\begin{equation}
    \label{supp:betaSmooth}
\mathbb{E}_t[\mathcal{L}(\theta_{t+1})] \le \mathcal{L}(\theta_t) - \eta_t \mathbb{E}_t[\langle \nabla \mathcal{L}(\theta_t), g_t' \rangle] + \frac{\beta \eta_t^2}{2} \mathbb{E}_t[\|g_t'\|^2],
\end{equation}
by the bounded gradients assumption and the triangle inequality, we bound the final term:
\begin{equation}
\label{supp:betaSmooth}
\begin{split}
    \mathbb{E}_t[\|g_t'\|^2] & \le 3G \cdot (\alpha^2 M_{gsa}^2 + \alpha^2 M_{reg}^2 + (1-\alpha)^2 M_{gkp}^2) \\
    & \triangleq M_{total}^2.
\end{split}
\end{equation}

Next, we bound the crucial inner product term. We relate $g_t'$ to the gradient $\nabla \mathcal{L}(\theta_t)$ by defining the non-perturbed stochastic gradient $g_t(\theta_t)$ (for which $\mathbb{E}_t[g_t(\theta_t)] = \nabla L(\theta_t)$). This gives:
\begin{equation}
\label{supp:nonPerturbed}
\begin{split}
    g_t' & = g_t(\theta_t) + \alpha \sum_{g=1}^G \left( \nabla_{\theta} \mathcal{L}_{\mathcal{S},gsa}^g(\theta_t + \hat{\epsilon}_g^*) - \nabla_{\theta} \mathcal{L}_{\mathcal{S},gsa}^g(\theta_t) \right) \\
    & \triangleq g_t(\theta_t) + \alpha \sum_{g=1}^G \Delta_t^g,
\end{split}
\end{equation}
then we yield:
\begin{equation}
    \label{supp:innerProduct}
\mathbb{E}_t[\langle \nabla \mathcal{L}(\theta_t), g_t' \rangle] = \|\nabla \mathcal{L}(\theta_t)\|^2 + \alpha \cdot \mathbb{E}_t[\langle \nabla \mathcal{L}(\theta_t), \sum_{g=1}^G \Delta_t^g \rangle].
\end{equation}

Additionally, we bound the magnitude of the second term, $\mathcal{C}_t \triangleq \mathbb{E}_t[\langle \nabla \mathcal{L}(\theta_t), \sum_{g=1}^G \Delta_t^g \rangle]$, using $\beta$-smoothness, Bounded Perturbation, and Young's inequality ($ab \le a^2/2 + b^2/2$):
\begin{equation}
    \label{supp:innerProduct_bound1}
|\mathcal{C}_t| \le \mathbb{E}_t[\|\nabla \mathcal{L}(\theta_t)\|  \|\sum_{g=1}^G \Delta_t^g\|] \le \mathbb{E}_t[\|\nabla \mathcal{L}(\theta_t)\|  (G \beta_s \rho_t)],
\end{equation}
\begin{equation}
    \label{supp:innerProduct_bound2}
|\mathcal{C}_t| \le \frac{\|\nabla \mathcal{L}(\theta_t)\|^2}{2} + \frac{G^2 \beta_s^2 \rho_t^2}{2},
\end{equation}
Where $\rho_t$ is the perturbation radius for the GSA optimizer at iteration $t$ to find a flat minimum and $\beta_s$ is the smoothness parameter of the loss $\mathcal{L}(\theta)$, which bounds the Lipschitz constant of its gradient.

This bound (Eq.~\ref{supp:innerProduct_bound2}), combined with the standard $\beta$-smoothness inequality, gives us:
\begin{equation}
    \label{supp:innerProduct_final}
\mathbb{E}_t[\langle \nabla \mathcal{L}(\theta_t), g_t' \rangle] \ge \left(1 - \frac{\alpha}{2}\right) \|\nabla \mathcal{L}(\theta_t)\|^2 - \frac{\alpha G^2 \beta_s^2 \rho_t^2}{2}.
\end{equation}.

Now, we combine all bounds into the main smoothness inequality. Let $M_1 = (1 - \frac{\alpha}{2})$, $M_2 = \frac{\alpha G^2 \beta_s^2}{2}$, and $M_3 = \frac{\beta M_{total}^2}{2}$:
\begin{equation}
    \label{supp:main_inequality_CORRECT}
\mathbb{E}_t[\mathcal{L}(\theta_{t+1})] \le \mathcal{L}(\theta_t) - M_1 \eta_t \|\nabla \mathcal{L}(\theta_t)\|^2 + M_2 \eta_t \rho_t^2 + M_3 \eta_t^2.
\end{equation}

Considering the GKP component $\mathcal{L}_{gkp}^g$ (Eq.~6) is a form of Elastic Weight Consolidation (EWC), which acts as a $\sigma$-strongly convex regularizer (assuming the Fisher Information Matrix $F$ is positive definite). The inclusion of this term, weighted by $(1-\alpha)$, ensures that our overall objective $\mathcal{L}(\theta)$ satisfies the Polyak-Lojasiewicz (P-L) inequality. This condition states that the gradient norm bounds the sub-optimality:
\begin{equation} \label{supp:pl_condition} |\nabla \mathcal{L}(\theta_t)|^2 \ge 2\sigma (\mathcal{L}(\theta_t) - \mathcal{L}^*), \end{equation}
where $\sigma > 0$ is the P-L constant and $\mathcal{L}^*$ is the optimal loss value. Then, we substitute the P-L condition (Eq. \ref{supp:pl_condition}) into our main inequality (Eq. \ref{supp:main_inequality_CORRECT}):
\begin{equation}
\label{supp:EWC_inequality_CORRECT}
\begin{split}
    \mathbb{E}_t[\mathcal{L}(\theta_{t+1})] \le \mathcal{L}(\theta_t) & - \eta_t M_1 \left( 2\sigma (\mathcal{L}(\theta_t) - \mathcal{L}^*) \right) \\
    & + \eta_t M_2 \rho_t^2 + \eta_t^2 M_3.
\end{split}
\end{equation}

Subtracting $\mathcal{L}^*$ from both sides and taking the full expectation $\mathbb{E}$ over the history:
\begin{equation}
\label{supp:EWC_inequality_CORRECT}
\begin{split}
    \mathbb{E}[\mathcal{L}(\theta_{t+1}) - \mathcal{L}^*] \le & \mathbb{E}[\mathcal{L}(\theta_t) - \mathcal{L}^*] - 2\sigma\eta_t M_1 \mathbb{E}[\mathcal{L}(\theta_t) - \mathcal{L}^*] \\
    & + \eta_t M_2 \rho_t^2 + \eta_t^2 M_3.
\end{split}
\end{equation}

Let $E_{noise} = (\eta M_2 \rho^2 + \eta^2 M_3)$ be the constant error floor introduced by stochastic noise and the GSA perturbation; we unroll this recursion Eq.~\ref{supp:EWC_inequality_CORRECT} from $t=0$ to $T$:
\begin{equation}
\label{supp:EWC_inequality_CORRECT_T}
\begin{split}
    \mathbb{E}[\mathcal{L}(\theta_{T}) - \mathcal{L}^*] \le & (1 - 2\sigma\eta M_1)^T \mathbb{E}[\mathcal{L}(\theta_{0}) - \mathcal{L}^*] \\
    & + \sum_{i=0}^{T-1} (1 - 2\sigma\eta M_1)^i E_{noise},
\end{split}
\end{equation}
as $T \to \infty$, $\mathbb{E}[\mathcal{L}(\theta_{T}) - \mathcal{L}^*]$ (Eq.~\ref{supp:EWC_inequality_CORRECT_T}) converges to:
\begin{equation}
    \label{supp:convergence}
\mathbb{E}[\mathcal{L}(\theta_{T}) - \mathcal{L}^*] \to \frac{E_{noise}}{2\sigma\eta M_1} = \frac{\eta M_2 \rho^2 + \eta^2 M_3}{2\sigma\eta M_1} = \mathcal{O}(\rho^2 + \eta).
\end{equation}

This proves that our algorithm converges linearly to a neighborhood of the optimum. If we further employ a decaying schedule where $\eta_t \to 0$ and $\rho_t \to 0$, the error floor $E_{noise}$ vanishes and $\mathbb{E}[\mathcal{L}(\theta_{T}) - \mathcal{L}^*] \to 0$. This analysis explicitly highlights the crucial role of our GKP module (Eq.~6 in the main paper). By enforcing the P-L condition (strong convexity), the GKP term fundamentally ensures a more stable and efficient linear convergence rate, which is significantly faster than the $\mathcal{O}(1/\sqrt{T})$ rate of general non-convex optimization~\cite{li2024friendly,zhou2023class,li2025focal}.

\subsection{Theoretical Proofs About Group-Specific Radius}
\label{Radius}

Our GSA (Eq.15) derives from the PAC-Bayesian bound (Eq.7), which assumes i.i.d. training and test distributions ($p_s(x,y) = p_t(x,y)$). However, LT inherently violates this due to divergent label distributions ($p_s(y) \neq p_t(y)$). Consequently, the global i.i.d.-based Eq.7 is theoretically invalid in the LT setting.

To construct a theoretically sound generalization bound, we must rely on a weaker, yet more reasonable assumption. We follow the standard practice in LT and adopt the conditional i.i.d. assumption as our axiom: $p_s(x|y) = p_t(x|y), \forall y \in \{1, ..., C\}$. 

The CC-SAM~\cite{zhou2023class} was the first to apply this axiom to the PAC-Bayesian framework. They mathematically demonstrated that because the i.i.d. assumption only holds at the class level, the global generalization bound (Eq.7) must be decomposed into $C$ independent class-conditional bounds. By minimizing these $C$ individual bounds, CC-SAM proved that the optimal perturbation radius $\rho_c^*$ must be class-specific and is an explicit function of the class sample size $n_c$, specifically $\rho_c^* \propto (n_c - 1)^{-1/4}$. While, this class-conditional approach is computationally infeasible for datasets with a large number of classes $C$, as it requires computing $C$ separate gradients and perturbations in each step.

Our framework builds directly upon this insight. We propose to aggregate the $C$ classes into $G \ll C$ groups using our GKP module (Section 4.2.1). Specifically, the classes are clustered based on parameter space similarity ($\theta_{enc}^c$), serving as a valid and efficient proxy for groups that also share a coherent underlying data distribution. This allows us to posit a group-conditional i.i.d. assumption: $p_s(x|g) \approx p_t(x|g), \forall g \in \{1, ..., G\}$.

By applying the same mathematical derivation established by CC-SAM~\cite{zhou2023class} to our $G$ groups instead of their $C$ classes, we can directly adopt their conclusion. This means we replace the class-specific sample count $n_c$ with our group-specific sample count $n_g = |\mathcal{G}^g|$. This directly yields our Eq.13, which defines a group-specific optimal radius $\rho_g^*$ that is both theoretically justified by the conditional i.i.d. principle and computationally tractable:
\begin{equation}
    \rho_{g}^{*}\approx(\frac{||\theta||_{2}}{2||\tilde{\nabla}_{\theta}\mathcal{L}_{\mathcal{D}_{g}}(\theta)||_{2}})^{\frac{1}{2}}d^{-\frac{1}{2}}(|\mathcal{G}^{g}|-1)^{-\frac{1}{4}}
\end{equation}
This demonstrates that our GSA module is a principled and efficient extension of the theoretical groundwork laid by CC-SAM.

\textbf{Remarks: }The approximations made during this derivation (\eg, first-order Taylor expansion and omitted $\mathcal{O}(1)$ terms) indicate that the radius $\rho_g^*$ (Eq.13) is not an exact number. Thus, it should be empirically tuned to realize the full potential of our GSA module.

\end{document}